\documentclass[10pt,twocolumn,letterpaper]{article}

\usepackage{cvpr}              % To produce the CAMERA-READY version
\usepackage{multirow}
\usepackage{bm}
\usepackage{tabularx}
\definecolor{cvprblue}{rgb}{0.21,0.49,0.74}
\usepackage[pagebackref,breaklinks,colorlinks,allcolors=cvprblue]{hyperref}

\def\confName{CVPR}
\def\confYear{2027}

\title{CMC-WAM: Causal Mechanism Coupling for World Action Models}

\author{
Yao Zhou$^{1,2,3}$ \quad
Fengge Wu$^{1,2}$ \quad
Changwen Zheng$^{1,2}$\quad
Wenwen Qiang$^{1,2}$\thanks{Corresponding author: \texttt{qiangwenwen@iscas.ac.cn}} \\[5pt]
{\small $^{1}$National Key Laboratory of Space Integrated Information System,} {\small Institute of Software, Chinese Academy of Sciences, Beijing, China}\\[2pt]
{\small $^{2}$Institute of Software, Chinese Academy of Sciences, Beijing, China}\\
{\small $^{3}$University of Chinese Academy of Sciences, Beijing, China}
}
\begin{document}
\maketitle

% \section{Abstract}
\begin{abstract}
World Action Models (WAMs) jointly generate actions and predict future observations for robotic control. However, existing unified architectures do not explicitly organize the shared and role-specific information required by these processes or characterize their distinct conditioning structures. We propose CMC-WAM, which decomposes unified DiT representations into shared and role-specific factors, and employs independent routers and expert networks to model action generation and world evolution. It further introduces asymmetric coupling: candidate actions condition future world prediction (A$\rightarrow$W), while predicted future states guide action selection (W$\rightarrow$A). Experiments on RoboCasa and LIBERO demonstrate improved mean success rates over Cosmos-Predict2.5 with a modest increase in trainable parameters. Real-world experiments on a dual-arm platform further demonstrate its applicability to physical manipulation.

\end{abstract}

\section{Introduction}

%%%%%%%%%%%%%%%%%%%% 背景 %%%%%%%%%%%%%%%%%%%%
% World Action Models (WAMs) have recently emerged as a promising paradigm for robot learning by extending world modeling from environment prediction towards action-conditioned prediction and control. By learning latent representations of the environment and modeling how future states evolve under different actions, WAMs provide a unified framework for prediction, planning, and control \cite{ha2018world,wu2023daydreamer,zhou2606tau0}. Recent advances further integrate large-scale video pretraining and robot interaction data into world-action modeling. Representative approaches, including Unified Video Action Model, Unified World Models, Cosmos Policy, and LingBot-VA, demonstrate the potential of combining video generation, action modeling, and robot control within a single framework\cite{li2025unifiedvideoactionmodel,zhu2025unified,kim2026cosmos,li2026causal}.

World Action Models (WAMs) extend world modeling to action-conditioned future prediction and control, learning environment representations and action-dependent dynamics to support unified prediction and planning ~\cite{ha2018world,wu2023daydreamer,zhou2606tau0}. Recent methods combine large-scale video pretraining with robot interaction data. Representative approaches, including Unified Video Action Model~\cite{li2025unifiedvideoactionmodel}, Unified World Models~\cite{zhu2025unified}, Cosmos Policy~\cite{kim2026cosmos}, and LingBot-VA~\cite{li2026causal}, demonstrate the potential of integrating video generation, action modeling, and robot control within a unified framework. %\cite{li2025unifiedvideoactionmodel,zhu2025unified,kim2026cosmos,li2026causal}.

%%%%%%%%%%%%%%%%%%%% 挑战 %%%%%%%%%%%%%%%%%%%%
% 世界动作模型（World Action Models, WAMs）将动作生成与未来世界预测纳入统一建模框架，使机器人能够同时决定当前应采取的动作，并预测动作执行后可能产生的环境变化。这两个过程具有不同的生成目标和信息需求：动作生成关注任务目标的实现，世界预测关注动作引起的物理状态变化。同时，二者又紧密关联，动作决定了世界如何演化，而对未来状态的预测也能够反过来指导动作评估与选择。

% 然而，现有 WAM 通常通过统一生成架构、共享骨干网络或跨模块信息交互来连接动作生成与未来世界预测 \cite{li2025unifiedvideoactionmodel,zhu2025unified,kim2026cosmos,li2026causal}。这些设计促进了信息共享与联合学习，但主要建立了计算层面的依赖，并未显式刻画两种生成过程各自不同的结构依赖，也未明确如何区分、组织和协调二者的共享信息与特定信息。因此，如何在统一框架下显式建模这两种既有差异又相互关联的生成过程，仍是 WAM 面临的重要问题。
World Action Models integrate action generation and future world prediction into a unified modeling framework. This enables robots to determine what actions to take while anticipating the resulting changes in the environment. These two processes have distinct generative objectives and information requirements: action generation focuses on achieving task goals, whereas world prediction models the physical consequences of actions. Meanwhile, they are closely coupled: actions drive world evolution, while predicted future states can in turn guide action evaluation and selection.

However, existing WAMs typically connect action generation and future world prediction through unified generative architectures, shared backbones, or cross-module information exchange ~\cite{li2025unifiedvideoactionmodel,zhu2025unified,kim2026cosmos,li2026causal}. Such designs facilitate information sharing and joint learning, but primarily establish computational dependencies. They do not explicitly characterize the distinct structural dependencies of the two processes or clarify how their shared and role-specific information should be organized and coordinated. Therefore, explicitly modeling these two distinct yet coupled generative processes within a unified framework remains an important challenge for WAMs.
% Despite these advances, current WAMs still face challenges in complex closed-loop robotic manipulation, particularly in long-horizon and contact-rich tasks \cite{kim2026cosmos,guo2026vlaw}. For example, in plate insertion into a narrow rack, small deviations in the generated actions may lead to different physical outcomes, such as successful insertion or collision. This requires a robot not only to generate goal-directed actions but also to predict the resulting physical state transitions. Since action generation and world evolution have distinct modeling requirements yet remain tightly coupled through physical interactions, how to represent and coordinate these two generative processes remains a key challenge for WAMs. % This motivates a closer examination of the two generative processes underlying WAMs, particularly their distinct dependencies and the relationships between them.

%%%%%%%%%%%%%%%%%%%% 动机 %%%%%%%%%%%%%%%%%%%%
% First, 我们提出一个因果视角分析。。。。。。

To better characterize the structural dependencies between action generation and world evolution, we revisit robot-environment interactions from a causal perspective. We formalize this perspective using a Structural Causal Model (SCM). As illustrated in Fig.~\ref{fig:causal_analysis}, we construct a causal graph with four nodes: the current physical interaction state $S_t$, task goal $G$, executed action $A_t$, and subsequent physical state $S_{t+1}$. The edges $S_t \rightarrow A_t$ and $G \rightarrow A_t$ represent the direct causal dependencies of action selection. Similarly, $S_t \rightarrow S_{t+1}$ and $A_t \rightarrow S_{t+1}$ describe the dependencies of physical state transitions. These relationships reveal two distinct yet coupled structural mechanisms governing action selection and world evolution. They have different causal dependencies, share the current interaction state, and are connected through action execution.
% However, existing WAMs typically connect action generation and future world prediction through unified generative architectures, shared backbones, or cross-module information exchange\cite{li2025unifiedvideoactionmodel,zhu2025unified,kim2026cosmos,li2026causal}. Such designs support information sharing and joint learning. While these computational dependencies alone do not explicitly specify how the distinct structural mechanisms are represented or how their interactions are coordinated.

%%%%%%%%%%%%%%%%%%%% 方法 %%%%%%%%%%%%%%%%%%%%
% 基于上述因果分析，我们提出因果机制耦合世界动作模型（Causal Mechanism Coupled World--Action Model, CMC-WAM），将动作生成与世界演化建模为两种结构依赖不同且相互关联的生成机制。具体而言，CMC-WAM 从动作与世界两个分支中提取共享因素与私有因素，并通过跨分支一致性与冗余抑制约束，分别组织两种生成过程共同依赖的交互信息及各自所需的特有信息。在此基础上，模型根据两种生成机制不同的结构依赖，采用独立的机制路由器与专家网络，分别建模动作生成与世界演化。进一步地，CMC-WAM 引入非对称机制耦合，将候选动作作为世界预测的条件，并利用预测的未来状态评价候选动作，从而实现两种生成机制的协调。上述表示与机制通过动作生成和未来世界预测任务进行联合优化。
Based on our causal analysis, we propose Causal Mechanism Coupling for World Action Model (CMC-WAM), which models action generation and world evolution as two distinct but coupled generative mechanisms. Specifically, CMC-WAM decomposes action- and world-branch representations into shared and private factors. Cross-branch consistency and redundancy-suppression constraints encourage the shared factors to capture common interaction information while preserving role-specific information in the private factors. To model their distinct generative roles, CMC-WAM employs independent mechanism routers and expert networks for action generation and world evolution, respectively. Furthermore, it introduces asymmetric mechanism coupling: candidate actions condition future world prediction (A$\rightarrow$W), while predicted future states guide candidate selection (W$\rightarrow$A). The CMC modules are jointly optimized using action-generation and future-world-prediction objectives, together with factorization constraints.
% Based on the above causal analysis, we propose the Causal Mechanism Coupled World Action Model (CMC-WAM), which models action generation and world evolution as distinct but coupled generative mechanisms. Specifically, CMC-WAM decomposes action- and world-branch representations into shared and private factors, and organizes their common interaction information and role-specific information through cross-branch consistency and redundancy suppression constraints. Based on the distinct structural dependencies of the two mechanisms, CMC-WAM employs independent mechanism routers and expert networks to model action generation and world evolution, respectively. Furthermore, CMC-WAM introduces asymmetric mechanism coupling, which conditions world prediction on candidate actions and uses the predicted future states to evaluate these candidates, thereby coordinating the two generative mechanisms. The resulting representations and mechanisms are jointly optimized through action generation and future world prediction objectives.

%%%%%%%%%%%%%%%%%%%% 贡献 %%%%%%%%%%%%%%%%%%%%
% （1）因果视角下的 WAM 机制分析。 从因果视角分析 WAM 中动作生成与世界演化之间的关系，刻画两种生成机制不同的条件依赖与计算信息流，明确共享交互信息、角色特有因素及其非对称依赖关系在 WAM 建模中的作用，为 WAM 的结构化建模与协调提供新的视角。
% （2）提出因果机制耦合世界动作模型 CMC-WAM。 通过共享与私有因素分解、独立机制路由及专家网络，分别建模动作生成与世界演化，并引入非对称机制耦合，实现动作条件化的未来世界预测与基于未来状态价值的候选动作选择，从而建立面向闭环控制的协调决策机制。
% （3）系统验证 CMC-WAM 的有效性与实际适用性。 在 Cosmos-Predict2.5 上实例化 CMC-WAM，并在 RoboCasa 和 LIBERO 两个机器人操作基准上进行闭环评估，实验结果表明该方法在两个基准上均优于原始模型。进一步通过消融实验和效率分析验证核心设计的作用，并在真实双臂机器人平台的三项操作任务上验证其物理执行能力。
Our key contributions are summarized as follows:
(1) We provide a \textbf{causal perspective} on the relationship between action generation and world evolution in WAMs. We characterize their distinct conditioning dependencies and computational information flows, highlighting the roles of shared interaction information, role-specific factors, and asymmetric dependencies. This provides a new perspective for structured modeling and coordination in WAMs.

(2) We propose \textbf{CMC-WAM}, a mechanism-coupling framework that integrates shared and private factorization, independent mechanism routing, and expert networks. It separately models action generation and world evolution, while enabling action-conditioned future prediction and value-guided candidate selection for closed-loop control.

(3) We evaluate CMC-WAM on \textbf{RoboCasa and LIBERO} using Cosmos-Predict2.5 as the base model. Experiments achieve higher mean success rates on both benchmarks. Further ablation and efficiency studies examine the contributions of individual components and their computational costs. Real-world experiments on a dual-arm robotic platform across three manipulation tasks further demonstrate its applicability to physical robotic execution.

\section{Related Work}
% 1. 现有 WAM 如何联合两种生成过程 → 
% 2. 为什么需要关注两种生成过程的结构关系 → 
% 3. 现有结构化与因果机制研究提供了什么启发

% 近年来，具身机器人研究逐渐将未来视觉预测引入策略学习与决策，形成了不同的世界—动作建模范式。一类方法利用视频生成模型或世界模型预测未来观测与环境状态，并将预测结果用于辅助机器人规划与控制 \cite{guo2026ctrl,zhou2024robodreamer,liao2026genie,huang2026enerverse,bu2024closed}。另一类方法将未来状态的潜在表征融入策略学习，通过预测性表示为动作生成提供信息 \cite{hu2024video,zheng2025flarerobotlearningimplicit}。近期研究进一步探索在统一框架中联合建模动作与未来世界状态，通过共享潜在表示、联合生成或多流协同实现两类任务的信息交互 \cite{guo2411prediction,li2025unifiedvideoactionmodel,zhu2025unified,kim2026cosmos,bi2026motus,li2026causal,won2026dust,zhou2606tau0}。这些研究从未来视觉预测、潜在状态建模到视频—动作联合生成，逐步拓展了世界模型与机器人策略的结合方式。然而，动作生成与世界预测既存在共同的信息需求，也具有各自特有的建模需求。如何在联合建模过程中合理组织两类任务的共享信息与特有信息，以及它们之间的联系，仍值得进一步研究。
\subsection{World Models for Robotic Control}
\label{sec:related_wam}

Recent studies in embodied robotics have increasingly incorporated future visual prediction into policy learning and decision-making, leading to diverse paradigms for world-action modeling. One line of work leverages video generative models or world models to predict future observations and environment states, using these predictions to facilitate robot planning and control \cite{guo2026ctrl,zhou2024robodreamer,liao2026genie,huang2026enerverse,bu2024closed}. Another line integrates latent representations of future states into policy learning, providing predictive information for action generation \cite{hu2024video,zheng2025flarerobotlearningimplicit}. More recent studies explore jointly modeling actions and future world states within unified frameworks, enabling information exchange between the two tasks through shared latent representations, joint generation, or multi-stream interactions \cite{guo2411prediction,li2025unifiedvideoactionmodel,zhu2025unified,kim2026cosmos,bi2026motus,li2026causal,zhou2606tau0}. These studies have expanded the integration of world models and robotic policies, ranging from future visual prediction and latent-state modeling to joint video-action generation. However, action generation and world prediction have both shared information requirements and task-specific modeling needs. How to appropriately organize the shared and task-specific information required by the two tasks, as well as the relationships between them, within a unified modeling framework remains an open question.

% ### 2.2 Causal Reasoning and Modular Mechanisms

% 因果学习已被广泛探索用于强化学习、机器人控制及世界模型，通过识别任务相关的因果因素、刻画状态与动作之间的结构依赖，以及学习跨环境稳定的表示，提升模型的预测能力与泛化性能 \cite{zhang2020invariant,sontakke2021causal,wang2022cdl,lee2023scale}。近年来，相关研究进一步探索了可组合的因果动力学建模，使智能体能够利用已有的动态规律适应未见环境与任务 \cite{wang2025modeling}。这些工作表明，因果建模能够为理解环境动态、识别关键交互因素以及学习具有泛化能力的世界表示提供有效的研究视角。

% 独立因果机制（Independent Causal Mechanisms, ICM）原则进一步为复杂动态系统的结构化建模提供了理论依据。该原则认为，复杂系统可以由多个具有相对自治性的生成机制构成，各机制具有自身的生成规律，并通过明确的因果依赖共同决定系统行为 \cite{parascandolo2018learning,besserve2021independent}。机制层面的自治性使得局部机制可以在其他机制保持相对稳定的条件下独立变化，从而为局部干预、机制组合及分布变化下的外推提供基础 \cite{besserve2021independent,gresele2021ima}。这一思想也被用于动态系统建模：通过具有相对独立转移规律的动态模块及稀疏信息交互，实现不同动态过程的专门化，并提升模型对局部因素变化的泛化能力 \cite{goyal2021rim,wang2022cdl,wang2025modeling}。因果机制和因果动力学为理解 WAM 内部不同功能过程的结构关系提供依据，也为具有专门化分支和结构化交互的架构提供潜在指导。

\subsection{Causal Modeling and Mechanisms}
\label{sec:related_causal}

Causal learning has been explored in reinforcement learning, robot control, and world modeling to identify task-relevant causal factors, characterize structured dependencies between states and actions, and learn representations that generalize across environments \cite{zhang2020invariant,sontakke2021causal,wang2022causal}. Recent studies further investigate compositional causal dynamics, enabling agents to leverage learned dynamical regularities in unseen environments and tasks \cite{wang2025modeling}. These efforts highlight the potential of causal modeling for understanding environment dynamics, identifying relevant interaction factors, and learning generalizable world representations.

From a broader theoretical perspective, the principle of Independent Causal Mechanisms (ICM) suggests that complex data-generating processes can be decomposed into relatively autonomous mechanisms, each governed by its own generative regularities while remaining connected through causal dependencies \cite{scholkopf2021toward}. This modularity enables localized interventions and mechanism recombination, providing a basis for understanding how complex systems respond to changes in individual mechanisms and generalize under distribution shifts \cite{gresele2021independent}. This principle has also been extended to dynamical systems, where modular transition mechanisms, sparse interactions, and structured state-action dependencies facilitate the modeling of distinct dynamic processes and improve generalization to unseen environments \cite{goyal2019recurrent,wang2022causal,wang2025modeling}. These perspectives highlight the importance of considering underlying causal structures when modeling complex dynamical systems, and may provide useful guidance for WAM architecture design, particularly in exploring specialized branches with structured interactions.

\section{Problem Formulation and Analysis}

% 本节目的：
% 简要定义机器人操纵数据、潜在状态、潜在动力学模型和策略模型，
% 本节不限定潜在状态的具体形式，也不讨论模型的优化目标。

\subsection{Problem Formulation}
\label{sec:problem_formulation}

% ============================================================
% 第一段：【本段目的：定义机器人操纵数据。】
% ============================================================

% 第一句：【本句目的：定义机器人操纵数据集。】
% 中文正文：
% 给定包含 $N$ 条机器人操纵轨迹的数据集，
% 每条轨迹由语言任务指令、观测序列和动作序列组成。
\textbf{Robot Manipulation Data.} Given a robot manipulation dataset containing $N$ trajectories,
\begin{equation}
    \mathcal{D}
    =
    \left\{
    \left(
        l,
        o_{0:T},
        a_{0:T-1}
    \right)^{(i)}
    \right\}_{i=1}^{N},
    \label{eq:wam_dataset}
\end{equation}
where $l$ denotes the language instruction, $o_t$ denotes the robot
observation at timestep $t$, which may include visual observations and proprioceptive states, and $a_t$ denotes the corresponding robot
action.

% ============================================================
% 第二段：【本段目的：定义 WAM 的潜在世界表征，并说明其编码来源和随交互过程的演化。】
% ============================================================
% 第一句：【本句目的：定义 WAM 的潜在世界表征。】
% 中文正文：
% 记 $s_t$ 为 WAM 在时刻 $t$ 形成的潜在世界表征，
% 用于表征机器人、操作对象及周围环境的内部信息。
\textbf{Latent World Representation} We use $s_t\in\mathcal{S}$ to denote the latent world representation formed by the WAM at timestep $t$, which represents internal information about the robot, manipulated objects, and the surrounding environment.
% 第二句：【本句目的：从编码角度说明潜在世界表征的来源，但不限定具体编码结构。】
% 中文正文：
% 从编码角度看，$s_t$ 由 WAM 对当前可获得的信息进行内部表征而形成，
% 这些信息可以包括语言任务指令、当前或历史观测以及先前执行的动作。
From an encoding perspective, $s_t$ is formed by internally representing the information available to the WAM, which may include the language instruction, current or historical observations, and previously executed actions.
% 第三句：【本句目的：说明不同 WAM 可以采用不同形式实现该潜在世界表征。】
% 中文正文：
% 根据具体架构，$s_t$ 可以对应视觉或视频潜变量、Transformer 隐藏表征、
% 多模态 token 特征，或其他参与世界建模和动作生成的中间表示。
Depending on the architecture, $s_t$ may be instantiated as visual or video latents, multimodal token features, or other intermediate representations involved in world modeling and action generation.
% 第四句：【本句目的：概括潜在世界表征随交互过程的演化及其不同实现方式。】
% 中文正文：
% 随着机器人执行动作并获得新的环境反馈，潜在世界表征随交互过程不断演化；不同 WAM 可通过显式潜在动力学、递归或自回归更新、对新观测的重新编码，或这些机制的组合实现这一过程。
% As the robot executes actions and receives new environmental feedback, the latent world representation evolves over interaction. Different WAMs may realize this process through explicit latent dynamics, recurrent or autoregressive updates, re-encoding of new observations, or combinations of these mechanisms.

% We denote the future latent-state sequence over a horizon of $K$ as
% \begin{equation}
%     \mathbf{s}_{t+1:t+K}
%     =
%     \left(
%         s_{t+1},
%         s_{t+2},
%         \ldots,
%         s_{t+K}
%     \right).
%     \label{eq:future_latent_states}
% \end{equation}

% ============================================================
% 第三段：【本段目的：定义潜在动力学模型。】
% ============================================================

% 第一句：【本句目的：定义未来潜在状态序列。】
% 中文正文：
% 记未来 $K$ 个时刻的潜在状态序列为
% $\mathbf{s}_{t+1:t+K}$。
% We denote the future latent-state sequence over a horizon of $K$ as
% \begin{equation}
%     \mathbf{s}_{t+1:t+K}
%     =
%     \left(
%         s_{t+1},
%         s_{t+2},
%         \ldots,
%         s_{t+K}
%     \right).
%     \label{eq:future_latent_states}
% \end{equation}

% ============================================================
% 第三段：【本段目的：定义 WAM 的未来世界状态生成过程。】
% ============================================================
\begin{figure}[t]
    \centering
    \includegraphics[width=0.85\linewidth]{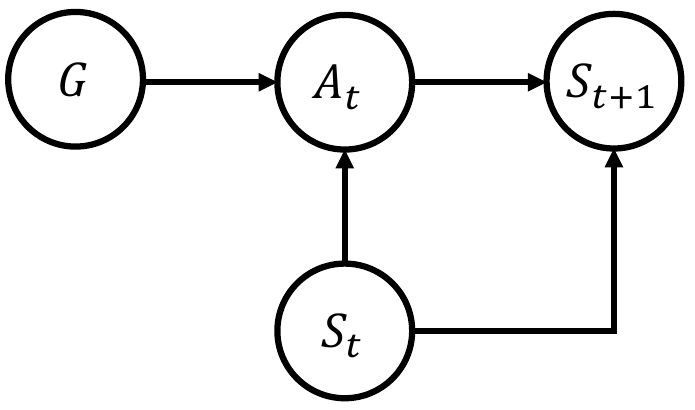}
    \caption{Structural causal graph of physical robot-environment interaction. Action selection $A_t$ depends on the current physical state $S_t$ and task goal $G$, while the subsequent physical state $S_{t+1}$ depends on $S_t$ and the executed action $A_t$.}
    \label{fig:causal_analysis}
\end{figure}
\textbf{Future World Prediction.}
Based on the available interaction information, a WAM
predicts the evolution of its latent world representation.
We denote the future latent-state sequence over a horizon
of $H$ as
\begin{equation}
    \mathbf{s}_{t+1:t+H}
    =
    \left(
        s_{t+1},
        s_{t+2},
        \ldots,
        s_{t+H}
    \right).
    \label{eq:future_latent_states}
\end{equation}
Depending on the architecture, future world prediction
may incorporate current or historical observations,
previous actions, and available action conditions,
without requiring a particular prediction architecture
or generation order.

% ============================================================
% 第三段：【本段目的：定义 WAM 的动作生成过程，并兼容不同动作生成形式。】
% ============================================================

% 第一句：【本句目的：说明动作生成依赖 WAM 当前形成的内部世界表征和任务信息。】
% 中文正文：
% 基于当前的潜在世界表征及任务信息，WAM 生成用于机器人与环境交互的控制动作。
\textbf{Action Generation} Based on the current latent world representation and task information, the WAM generates control actions for robot tasks.
% 第二句：【本句目的：定义动作块并兼容单步动作生成。】
% 中文正文：
% 对于采用动作块预测的 WAM，记从时刻 $t$ 开始、长度为 $H$ 的动作序列为
For WAMs that generate action chunks, we denote the action sequence of horizon $H$ starting from timestep $t$ as
$$
\mathbf{a}_{t:t+H-1}
=
\left(
a_t,
a_{t+1},
\ldots,
a_{t+H-1}
\right),
$$
where $H=1$ corresponds to single-step action generation.
% 第三句：【本句目的：说明不同 WAM 可以采用不同的动作生成实现，而不限定独立策略模型。】
% 中文正文：
% 根据具体架构，动作可以由独立的策略模块生成，也可以与未来世界表征联合生成，或通过统一的生成模型产生。
Depending on the architecture, actions may be generated by a dedicated policy module, jointly with future world representations, or through a unified generative model.

% ============================================================
% 第四段：【本段目的：区分 WAM 的训练目标与最终控制性能评估。】
% ============================================================

% 第一句：【本句目的：说明不同 WAM 采用架构相关的训练目标学习动作与未来世界信息。】
% 中文正文：
% 不同 WAM 可以采用不同的生成式学习目标，对机器人轨迹中的动作及未来世界信息进行建模，
% 其训练过程可统一表示为最小化模型特定的训练损失。
\textbf{Learning Objective and Control Evaluation.}
Different WAMs may employ different generative learning objectives to model robot actions and future world evolution from interaction trajectories. Their training process can be generally expressed as
\begin{equation}
    \Theta^{*}
    =
    \arg\min_{\Theta}
    \mathcal{L}_{\mathrm{WAM}}
    \left(
        \Theta;\mathcal{D}
    \right),
    \label{eq:wam_training_objective}
\end{equation}
where $\mathcal{L}_{\mathrm{WAM}}$ denotes the architecture-specific training objective, such as flow-matching, diffusion-based, or autoregressive prediction losses.

% 第二句：【本句目的：说明任务成功率用于衡量训练后 WAM 的最终机器人控制能力，而非作为统一训练目标。】
% 中文正文：
% 训练完成后，WAM 的最终机器人控制能力通过其在环境中执行语言指定任务的成功情况进行评估。
After training, the downstream control capability of a WAM is evaluated by whether its generated actions successfully accomplish the instructed manipulation task during environment interaction.
% 第三句：【本句目的：定义任务成功率作为本文主要控制性能指标。】
% 中文正文：
% 记 $\tau_{\Theta}$ 为参数为 $\Theta$ 的 WAM 执行得到的机器人轨迹，
% 并以任务成功率作为本文衡量下游控制性能的主要指标。
% Let $\tau_{\Theta}$ denote the trajectory induced by a WAM parameterized by $\Theta$, and let $\mathcal{S}(\tau_{\Theta},l)\in\{0,1\}$ indicate whether the trajectory successfully completes the task specified by instruction $l$.
Accordingly, we use task success rate (SR) as the primary metric for evaluating the downstream control performance of WAMs.
\begin{figure}[t]
    \centering
    \includegraphics[width=0.9\linewidth]{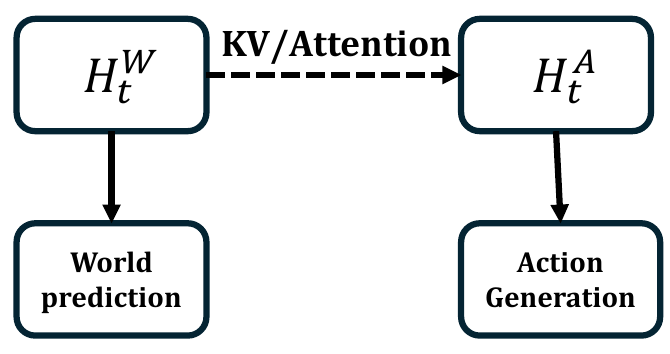}
    \caption{Illustrative neural computation graph of a WAM. World representations $H_t^W$ provide information to action representations $H_t^A$ through KV-cache sharing or attention. These computational dependencies describe information flow rather than physical causal relationships.}
    \label{fig:computation}
\end{figure}
\subsection{Causal Analysis}
\label{sec:causal_analysis}

% ============================================================
% Paragraph 1: 为什么要研究真实交互结构
% ============================================================

% WAM 在潜在空间中学习动作生成与未来世界预测，其生成的动作最终需要作用于真实环境，而动作执行后的状态变化受到物理世界演化规律的约束。为了明确动作生成与世界演化各自遵循什么规律，以及二者如何相互关联，我们首先从因果视角分析机器人与环境的交互过程。
WAM learns to generate actions and predict future world states within a latent space. Since the generated actions are ultimately applied to a real-world environment, the resulting state changes are constrained by the laws governing the evolution of the physical world. To clarify the distinct principles underlying action generation and world evolution—as well as how these two processes are interrelated, we begin by analyzing the interaction between a robot and its environment from a causal perspective.

% ============================================================
% Paragraph 2: SCM及其结构方程
% ============================================================

% 具体而言，我们考虑以下结构因果模型（Structural Causal Model, SCM）：

% [
% \begin{aligned}
% A_t &= f_A(S_t,G),\
% S_{t+1} &= f_W(S_t,A_t).
% \end{aligned}
% ]
% 其中，(S_t) 和 (S_{t+1}) 分别表示动作执行前后的任务相关物理交互状态，(G) 表示任务目标，(A_t) 表示实际执行的动作。$f_A$ 和 $f_W$ 分别表示动作选择与世界状态转移的结构函数。这里的 (S_t) 区别于 WAM 根据观测形成的潜在表征 (s_t)。如图 X(a) 所示，动作选择受到当前状态与任务目标的影响，而世界状态转移受到当前状态及实际执行动作的影响。
Specifically, we consider the following SCM of robot-environment interaction:
\begin{equation}
\begin{aligned}
A_t &= f_A(S_t,G),\\
S_{t+1} &= f_W(S_t,A_t).
\end{aligned}
\label{eq:interaction_scm}
\end{equation}
Here, $S_t$ and $S_{t+1}$ denote the physical interaction states before and after action execution, respectively. $G$ represents the task goal and $A_t$ is the action executed by the robot. The structural functions $f_A$ and $f_W$ characterize the action selection and physical state transition, respectively. Note that $S_t$ represents the underlying physical interaction state, which is conceptually distinct from the latent representation $s_t$ inferred by the WAM from observations. As illustrated in Fig.~\ref{fig:causal_analysis}, action selection depends on the current state and task goal, whereas physical state transitions depend on the current state and the executed action.
\begin{figure*}[t]
    \centering
    \includegraphics[width=\textwidth]{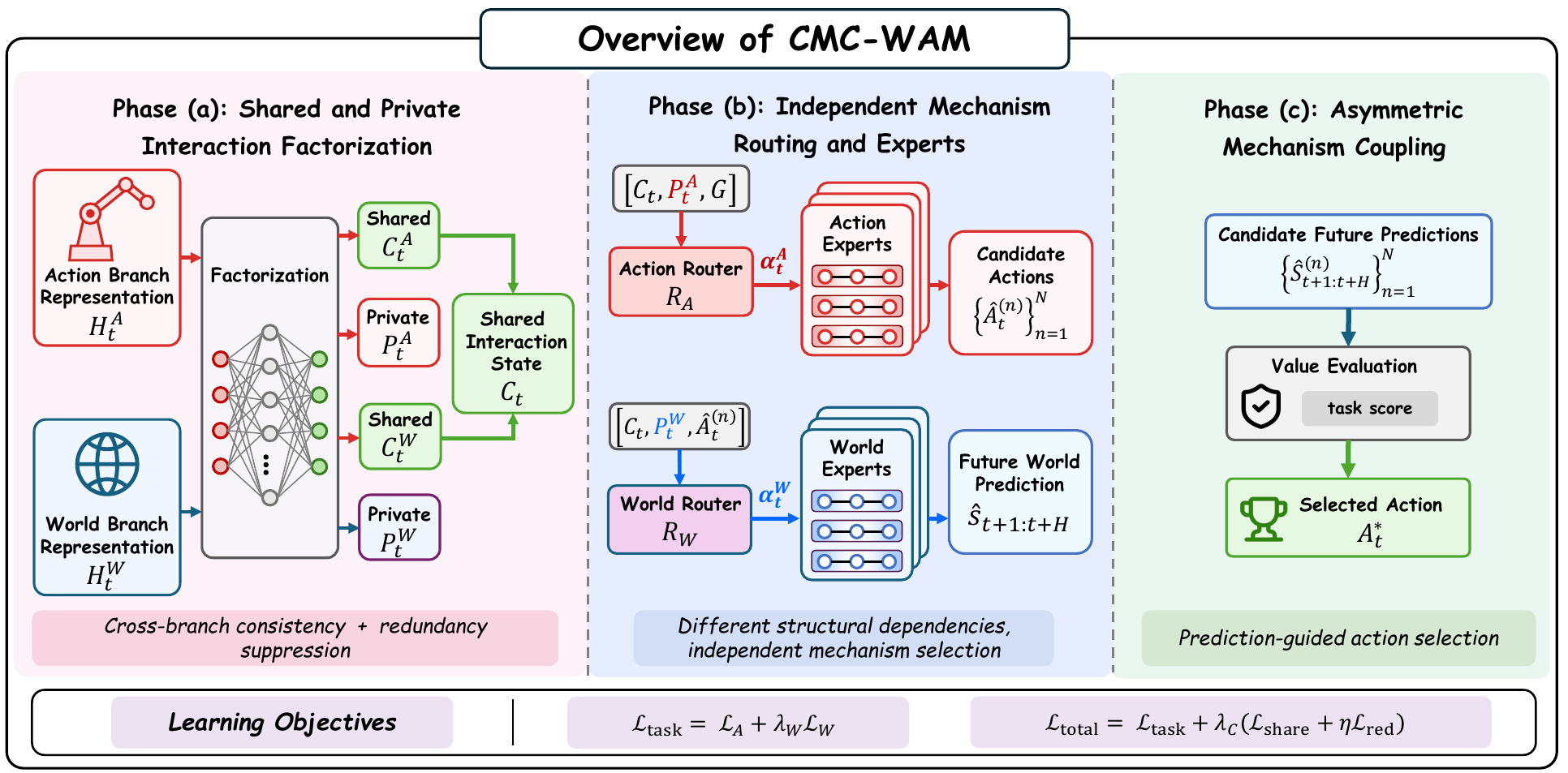}
    \caption{
    \textbf{Overview of the proposed CMC-WAM.}
    The framework consists of three components: (a) shared and private interaction factorization, (b) independent mechanism routing and experts for action and world generation, and (c) asymmetric mechanism coupling for prediction-guided action selection.
    }
    \label{fig:cmc_wam_architecture}
\end{figure*}
% ============================================================
% Paragraph 3: 从SCM推导两种不同但关联的机制
% ============================================================

% 从因果视角来看，上述两个结构方程分别描述了动作选择与物理状态转移的生成规律，对应两种不同的结构机制。二者具有不同的因果依赖：任务目标影响机器人如何选择动作，而实际执行动作直接影响后续物理状态。与此同时，两种机制通过当前交互状态和实际动作形成联系。由此，动作生成与状态演化具有不同但相互关联的结构机制。这为 WAM 如何组织潜在信息并协调动作生成与未来世界预测提供了
%From a causal perspective, the two structural equations characterize distinct mechanisms underlying action selection and physical state transitions. The task goal influences action selection, while the executed action directly affects the subsequent physical state. Meanwhile, both mechanisms depend on the current interaction state and are coupled through action execution. Thus, robot-environment interaction involves two distinct yet coupled structural mechanisms governing action selection and physical world evolution. This provides guidance for organizing latent information and coordinating action generation with future world prediction.

From a causal perspective, the two structural equations describe action selection and physical state transitions as distinct but coupled mechanisms. The task goal influences action selection, while the executed action directly affects the subsequent physical state; both mechanisms depend on the current interaction state and are linked through action execution. This distinction motivates organizing shared and role-specific latent information and coordinating action generation with future-world prediction.

% ============================================================
% （弃用）Paragraph 4: 重新审视现有WAM。
% ============================================================
% 上述因果分析为 WAM 的建模提供了结构参照。尽管 WAM 的潜在表示与预测状态不等同于真实物理状态，其建模仍需关注动作选择与物理世界演化所依赖的结构信息，包括两种生成过程的共同依赖、各自的角色差异及其相互联系。这为 WAM 如何组织潜在信息并协调动作生成与未来世界预测提供了指导。
% This provides structural guidance for WAM modeling. Although latent representations and predicted future states are distinct from underlying physical states, WAM modeling should account for the structural dependencies underlying action selection and physical world evolution, including their shared dependencies, role-specific requirements, and interconnections.

% ============================================================
% Paragraph 4: 重新审视现有 WAM 的计算结构及其局限。
% ============================================================

% 基于上述因果分析，我们进一步审视现有 WAM 如何表达两种生成过程及其关联。现有方法采用统一生成、共享表征，以及 KV cache、cross-attention 等跨模块信息交互方式，使动作生成与未来世界预测能够共享或交换信息。然而，这些设计主要从计算层面建立两个生成过程之间的联系，而计算依赖本身并不能揭示模型如何表达各自不同的结构依赖，以及如何利用和协调二者的共同信息。如图2所示，World 表征通过 KV cache 或 Attention 向 Action 表征传递信息，体现的是模型内部的计算依赖，并不能直接视为真实交互中的物理因果关系。尽管现有 WAM 可以隐式学习这些关系，但如何有效表达两种生成机制各自的结构依赖，并协调二者之间的信息交互，仍是一个值得深入研究的问题。

Building on this analysis, we examine how existing WAMs represent and connect action generation and future world prediction. Prior methods use joint video-action representations or unified Transformer backbones, with cross-modal attention exchanging information and KV caches to retain interaction history \cite{li2025unifiedvideoactionmodel,zhu2025unified,kim2026cosmos,li2026causal}. These designs establish computational dependencies, but do not explicitly characterize the distinct structural dependencies of the two generative processes or how their shared information should be coordinated. As illustrated in Fig.~\ref{fig:computation}, information transfer from world representations $H_t^W$ to action representations $H_t^A$ reflects an internal computational dependency rather than a physical causal relationship. Although existing WAMs may implicitly capture aspects of these relationships, how to explicitly represent and coordinate the two generative mechanisms remains unresolved.

The SCM analysis reveals that action generation and world evolution share the current interaction state but have distinct conditioning variables. This motivates CMC-WAM to organize their representations into shared and role-specific factors and coordinate the two generative processes accordingly.

\section{Method}

% CMC-WAM 从结构因果机制的角度组织动作生成与世界演化过程。首先，分别从两个生成分支中提取共享因素与私有因素，并融合共享信息形成共同交互状态。其次，根据两种生成机制不同的结构依赖，构建独立的机制路由器与专家网络，分别建模动作生成与世界演化。进一步地，通过非对称机制耦合，将候选动作作为世界预测的条件，并利用预测的未来状态指导动作选择，实现两种生成机制的协调。最后，通过生成任务监督与因素分解约束，联合优化共享与私有表示及两种生成机制。
This section presents CMC-WAM, which models action generation and world evolution as two structurally distinct yet coupled generative mechanisms. As illustrated in Fig.~\ref{fig:cmc_wam_architecture}, CMC-WAM first decomposes the representations of the two branches into shared and role-specific private factors, capturing common interaction information and mechanism-specific dependencies. It then employs independent mechanism routers and expert networks to model the two generative processes. To coordinate them, CMC-WAM introduces asymmetric coupling: candidate actions condition future world prediction (A$\rightarrow$W), while predicted future states guide action selection (W$\rightarrow$A). Finally, the resulting representations and mechanisms are jointly optimized through task supervision and factorization constraints.

% 提取公共因素/私有因素
% 1. 定义两类分支表示，以及共享因素与私有因素
% 2. 融合两分支的共享因素，构建共同交互状态
% 3. 私有因素
\subsection{Shared and Private Interaction Factorization}
% 动作生成与未来世界预测具有不同的生成目标和结构依赖，因此，两种生成过程需要的信息既存在共同部分，也具有各自的特有部分。为此，CMC-WAM 分别从动作分支和世界分支的相关表示中提取共享因素与私有因素。记两个分支经过相应编码器得到的隐藏表示为 \(H_t^A\) 和 \(H_t^W\)，通过角色特定的投影头得到：
% \[
% \begin{aligned}
% C_t^A &=P_C^A(H_t^A),&
% P_t^A &=P_P^A(H_t^A),\\
% C_t^W &=P_C^W(H_t^W),&
% P_t^W &=P_P^W(H_t^W).
% \end{aligned}
% \]
% 其中，\(C_t^A\) 和 \(C_t^W\) 分别表示从两个分支提取的共享因素，\(P_t^A\) 和 \(P_t^W\) 则表示动作生成与世界预测各自的私有因素。两个分支可以保留不同的输入形式与表示结构，无需强制其原始特征完全一致。

% 为表达两种生成过程共同依赖的交互信息，我们进一步融合两个分支的共享因素，构建统一的共享交互状态：
% \[
% C_t=F_C(C_t^A,C_t^W),
% \]
% 其中，\(F_C\) 表示共享信息融合函数，\(C_t\) 用于承载两种生成过程共同需要的交互信息，例如机器人与操作对象之间的相对关系、接触状态及相对运动等。考虑到两个分支提取的共享因素并不天然具有一致的语义，我们通过跨分支一致性约束促进其共同信息表达，并结合共享表示容量限制，避免共享通道过度吸收角色特有信息。

% 与此同时，CMC-WAM 保留两个分支的私有因素，以表达各自生成任务所需的特有信息。其中，动作私有因素 \(P_t^A\) 服务于动作机制选择与动作生成，世界私有因素 \(P_t^W\) 则参与世界机制选择与未来状态预测。为减少共享与私有表示之间的信息冗余，我们进一步引入冗余抑制约束，并利用各自的生成任务监督促进私有因素学习相应的任务相关信息。上述约束共同作用于共享与私有因素的学习，其具体优化形式将在第 4.4 节介绍。最终，模型形成共享交互状态 \(C_t\) 以及分别服务于动作生成和世界预测的私有因素 \(P_t^A\) 与 \(P_t^W\)，为后续的独立机制路由与专家建模提供结构化的信息基础。

To instantiate this structure, CMC-WAM extracts shared and private factors from the respective representations of the action and world branches. Let $H_t^A$ and $H_t^W$ denote the hidden representations. Role-specific projection heads are applied to obtain:
\begin{equation}
\begin{aligned}
C_t^A &= P_C^A(H_t^A), &
P_t^A &= P_P^A(H_t^A),\\
C_t^W &= P_C^W(H_t^W), &
P_t^W &= P_P^W(H_t^W).
\end{aligned}
\end{equation}
Here, $C_t^A$ and $C_t^W$ denote the shared factors extracted from the two branches, while $P_t^A$ and $P_t^W$ represent the private factors specific to action generation and world prediction, respectively. The two branches can retain different input modalities and representation structures without requiring their original features to be directly aligned.

To capture the interaction information commonly required by both generative processes, we further fuse the shared factors from the two branches into a unified shared interaction state:
\begin{equation}
C_t = F_C(C_t^A,C_t^W),
\end{equation}
where $F_C$ denotes the shared information fusion function. The resulting $C_t$ captures interaction information relevant to both processes, such as the relative configuration between the robot and manipulated objects, contact states, and relative motion. Since the shared factors extracted from the two branches do not necessarily have consistent semantics, we employ cross-branch consistency constraints to encourage their common information to align. We further restrict the dimensionality of the shared projections to limit the capacity of the common representation, reducing the tendency to absorb role-specific information.

Meanwhile, CMC-WAM retains the private factors to preserve the information specific to each generative process. Specifically, $P_t^A$ supports action mechanism selection and action generation, whereas $P_t^W$ contributes to world mechanism selection and future state prediction. To reduce information redundancy between shared and private representations, we introduce a redundancy suppression constraint. Meanwhile, the respective generative task objectives encourage the private factors to retain task-relevant information. The corresponding constraints are jointly applied to the learning of shared and private factors, with their optimization details provided in Section~4.4. Consequently, CMC-WAM obtains a shared interaction state $C_t$ and two role-specific private factors, $P_t^A$ and $P_t^W$, providing structured representations for subsequent independent mechanism routing and expert modeling.

% 机制与路由专家
% 1. 定义独立机制路由
% 2. 独立选择对应的机制专家
% 3. 专家组合形成不同的生成过程
\subsection{Independent Mechanism Routing and Experts}
% 在获得共享交互状态与私有因素后，CMC-WAM 进一步根据两种生成过程不同的结构依赖，分别建模动作生成与世界演化机制。具体而言，动作机制依赖共享交互状态、动作私有因素及任务目标，而世界机制则依赖共享交互状态、世界私有因素及动作条件。基于这一差异，我们为两个分支分别构建独立的机制路由器，其选择权重定义为：
% [
% \alpha_t^A=\operatorname{softmax}\left(\frac{R_A([C_t,P_t^A,e_G])}{\tau_A}\right),
% \qquad
% \alpha_t^W=\operatorname{softmax}\left(\frac{R_W([C_t,P_t^W,e(A_t)])}{\tau_W}\right).
% ]
% 其中，(R_A) 与 (R_W) 分别表示动作与世界机制路由器，(e_G) 和 (e(A_t)) 表示任务目标与动作条件的嵌入表示，(\tau_A) 和 (\tau_W) 为路由温度。两套路由器共享交互信息基础，但具有不同的条件输入与独立参数，因此无需强制其机制选择结果保持一致。

% 为刻画不同交互条件下的生成规律，我们分别构建动作专家集合 (\mathcal E_A={E_j^A}{j=1}^{K_A}) 和世界专家集合 (\mathcal E_W={E_k^W}{k=1}^{K_W})，其中 (K_A) 和 (K_W) 分别为两类专家的数量。各专家通过独立参数化的网络学习相应的生成映射，路由器则根据当前交互条件动态组合其输出。以 Flow Matching 为例，两个生成分支的速度场可以表示为：
% [
% \begin{aligned}
% v_\theta^A(z_\tau^A,\tau)
% &=\sum_{j=1}^{K_A}\alpha_{t,j}^A
% v_j^A(z_\tau^A,\tau\mid C_t,P_t^A,G),\
% v_\theta^W(z_\tau^W,\tau)
% &=\sum_{k=1}^{K_W}\alpha_{t,k}^W
% v_k^W(z_\tau^W,\tau\mid C_t,P_t^W,A_t).
% \end{aligned}
% ]
% 其中，(z_\tau^A) 和 (z_\tau^W) 分别表示动作与世界生成过程在流时间 (\tau) 的中间变量，(v_j^A) 和 (v_k^W) 表示相应专家的速度场。通过对专家输出进行加权组合，两种生成过程能够根据当前交互条件形成不同的生成映射，并分别完成动作生成与未来世界预测。专家的功能分工由生成任务监督与机制结构约束共同学习，具体优化目标将在第 4.4 节介绍。

Given the shared interaction state and private factors, CMC-WAM models action generation and world evolution as two distinct mechanisms according to their respective structural dependencies. Specifically, the action mechanism depends on the shared interaction state, action-specific private factors, and task goal, whereas the world mechanism depends on the shared interaction state, world-specific private factors, and action condition. Accordingly, we construct independent mechanism routers for the two branches, with their routing weights defined as:
\begin{equation}
\begin{aligned}
\alpha_t^A &=
\operatorname{softmax}\left(
\frac{R_A([C_t,P_t^A,e_G])}{\tau_A}
\right),\\
\alpha_t^W &=
\operatorname{softmax}\left(
\frac{R_W([C_t,P_t^W,e(A_t)])}{\tau_W}
\right).
\end{aligned}
\end{equation}
Here, $R_A$ and $R_W$ denote the action and world mechanism routers, respectively. $e_G$ and $e(A_t)$ represent the embeddings of the task goal and action condition, while $\tau_A$ and $\tau_W$ are the corresponding routing temperatures. Although both routers share the same interaction information, they have distinct conditional inputs and independent parameters. Therefore, their mechanism selection does not need to be explicitly aligned.

To characterize different generative patterns, we construct separate expert sets for action generation and world evolution, denoted by $\mathcal{E}_A=\{E_j^A\}_{j=1}^{K_A}$ and $\mathcal{E}_W=\{E_k^W\}_{k=1}^{K_W}$, respectively, where $K_A$ and $K_W$ are the numbers of experts in the two branches. Each expert is independently parameterized to learn mechanism-conditioned corrections to the pretrained generative velocity field. The routers dynamically combine these corrections according to the current interaction conditions. Taking Flow Matching as an example, the velocity fields of the two generative branches are formulated as:
% \begin{equation}
% \begin{aligned}
% v_\theta^A(z_\tau^A,\tau)
% &=
% v_{\mathrm{base}}^A(z_\tau^A,\tau)
% +
% \sum_{j=1}^{K_A}\bar{\alpha}_{t,j}^A
% \Delta v_j^A(z_\tau^A,\tau\mid C_t,P_t^A,G),\\
% v_\theta^W(z_\tau^W,\tau)
% &=
% v_{\mathrm{base}}^W(z_\tau^W,\tau)
% +
% \sum_{k=1}^{K_W}\bar{\alpha}_{t,k}^W
% \Delta v_k^W(z_\tau^W,\tau\mid C_t,P_t^W,A_t).
% \end{aligned}
% \end{equation}

\begin{equation}
\begin{aligned}
v_\theta^A(z_\tau^A,\tau)
&= v_{\mathrm{base}}^A(z_\tau^A,\tau)\\
&\quad + \sum_{j=1}^{K_A}
\bar{\alpha}_{t,j}^A
\Delta v_j^A(z_\tau^A,\tau
\mid C_t,P_t^A,G),\\[3pt]
v_\theta^W(z_\tau^W,\tau)
&= v_{\mathrm{base}}^W(z_\tau^W,\tau)\\
&\quad + \sum_{k=1}^{K_W}
\bar{\alpha}_{t,k}^W
\Delta v_k^W(z_\tau^W,\tau
\mid C_t,P_t^W,A_t).
\end{aligned}
\end{equation}
Here, $A_t$ denotes the action sequence conditioning the World branch. During training, it corresponds to the action sequence associated with the observed transition. During candidate evaluation, it is instantiated as the corresponding candidate $\hat{A}_t^{(n)}$, enabling the World branch to predict candidate-specific future states.
Moreover, $z_\tau^A$ and $z_\tau^W$ denote the intermediate variables of action generation and world prediction at flow time $\tau$, respectively. $v_{\mathrm{base}}^A$ and $v_{\mathrm{base}}^W$ are the velocity fields produced by the frozen pretrained WAM, while $\Delta v_j^A$ and $\Delta v_k^W$ denote the corrections predicted by the corresponding experts. The sparse routing weights $\bar{\alpha}_{t,j}^A$ and $\bar{\alpha}_{t,k}^W$ are normalized over the selected experts. By combining these expert corrections with the pretrained velocity fields, the two branches adapt their generative dynamics to the current interaction context while retaining the original model predictions.

% To characterize different generative patterns, we construct separate expert sets for action generation and world evolution, denoted by $\mathcal{E}_A=\{E_j^A\}_{j=1}^{K_A}$ and $\mathcal{E}_W=\{E_k^W\}_{k=1}^{K_W}$, respectively, where $K_A$ and $K_W$ are the numbers of experts in the two branches. Each expert is parameterized independently to learn a corresponding generative mapping, while the routers dynamically combine their outputs according to the current interaction conditions. Taking Flow Matching as an example, the velocity fields of the two generative branches are formulated as:
% \begin{equation}
% \begin{aligned}
% v_\theta^A(z_\tau^A,\tau)
% &=
% \sum_{j=1}^{K_A}\alpha_{t,j}^A
% v_j^A(z_\tau^A,\tau\mid C_t,P_t^A,G),\\
% v_\theta^W(z_\tau^W,\tau)
% &=
% \sum_{k=1}^{K_W}\alpha_{t,k}^W
% v_k^W(z_\tau^W,\tau\mid C_t,P_t^W,A_t).
% \end{aligned}
% \end{equation}
% Here, $z_\tau^A$ and $z_\tau^W$ denote the intermediate variables of action generation and world prediction at flow time $\tau$, respectively. $v_j^A$ and $v_k^W$ represent the velocity fields of the corresponding experts. By combining expert outputs through routing weights, the two branches form distinct generative mappings conditioned on the current interaction context, enabling action generation and future world prediction, respectively.

% 机制协调
\subsection{Asymmetric Mechanism Coupling}
% 尽管前述因素分解与独立机制路由能够分别建模动作生成和世界演化，两种生成机制仍需通过交互形成面向任务目标的决策。根据前述因果分析，动作生成与世界演化具有不同的结构依赖，并通过动作执行建立联系。基于此，CMC-WAM 引入非对称机制耦合：动作生成机制提供候选动作及其执行条件，世界演化机制预测相应的未来状态，并将预测结果反馈至动作选择。

% 具体而言，给定当前交互状态与任务目标，动作生成机制产生 (N) 个候选动作序列 ({\widetilde A_t^{(n)}}_{n=1}^{N})。对于每个候选动作，世界机制分别计算相应的路由权重，并预测其对应的未来状态：

% [
% \begin{aligned}
% \alpha_t^{W,(n)}
% &=R_W(C_t,P_t^W,\widetilde A_t^{(n)}),\
% \hat{s}_{t+1+H}^{(n)}
% &=F_W(s_t,\widetilde A_t^{(n)};\alpha_t^{W,(n)}).
% \end{aligned}
% ]

% 其中，(\alpha_t^{W,(n)}) 表示候选动作对应的世界机制选择权重，(\hat{s}_{t+1+H}^{(n)}) 表示预测的未来状态序列。随后，利用任务相关的价值与风险函数评价不同候选动作的预测结果，并选择最终执行的动作：

% [
% \begin{aligned}
% J^{(n)}
% &=V_G(\hat{s}^{(n)},G)
% -\beta,\operatorname{Risk}(\hat{s}^{(n)}),\
% n^&=\arg\max_n J^{(n)},
% \qquad A_t^=\widetilde A_t^{(n^*)}.
% \end{aligned}
% ]

% 由此，Action-to-World 耦合通过动作条件影响世界机制选择及未来预测，World-to-Action 耦合则通过预测结果评价影响动作决策。两种方向分别对应动作条件下的状态演化与模型内部的预测辅助决策，共同构成 CMC-WAM 的非对称机制协调过程。在闭环执行中，模型仅执行选中动作序列的首段，并根据新的观测持续更新机制表示与决策。

Although the aforementioned factorization and independent mechanism routing enable separate modeling of action generation and world evolution, the two generative mechanisms must interact to support task-oriented decision-making. As established by our causal analysis, these mechanisms exhibit distinct structural dependencies and are connected through action execution. Accordingly, CMC-WAM introduces asymmetric mechanism coupling: the action generation mechanism provides candidate actions and their execution conditions, while the world evolution mechanism predicts the corresponding future states and feeds these predictions back into action selection.

Specifically, given the current interaction state and task goal, the action generation mechanism produces $N$ candidate action sequences $\{\hat{A}_t^{(n)}\}_{n=1}^{N}$. For each candidate, the world mechanism computes the corresponding routing weights and predicts the future states:
% \begin{equation}
% \begin{aligned}
% \alpha_t^{W,(n)}
% &=
% \operatorname{softmax}\left(
% \frac{
% R_W([C_t,P_t^W,e(\widetilde{A}_t^{(n)})])
% }{\tau_W}
% \right),\\
% \hat{s}_{t+1:t+H}^{(n)}
% &=
% F_W(s_t,\widetilde{A}_t^{(n)};
% \alpha_t^{W,(n)}).
% \end{aligned}
% \end{equation}
\begin{equation}
\begin{aligned}
\alpha_t^{W,(n)}
&= \operatorname{softmax}\left(
\frac{R_W([C_t,P_t^W,e(\hat{A}_t^{(n)})])}{\tau_W}
\right),\\
\hat{s}_{t+1:t+H}^{(n)}
&= F_W(s_t,\hat{A}_t^{(n)};\alpha_t^{W,(n)}).
\end{aligned}
\end{equation}
Here, $\alpha_t^{W,(n)}$ denotes the world mechanism routing weights conditioned on the $n$-th candidate action, and $\hat{s}_{t+1:t+H}^{(n)}$ represents the corresponding predicted future state sequence. Subsequently, a task-related value function evaluates the predicted outcomes of different candidates, and the final action is selected as:
% \begin{equation}
% \begin{aligned}
% J^{(n)}
% &=
% V_G(\hat{s}^{(n)},G)
% -\beta\,\operatorname{Risk}(\hat{s}^{(n)}),\\
% n^*
% &=
% \arg\max_n J^{(n)},\\
% A_t^*
% &=
% \widetilde{A}_t^{(n^*)}.
% \end{aligned}
% \end{equation}
\begin{equation}
\begin{aligned}
J^{(n)} &= V_G(\hat{s}_{t+H}^{(n)},G),\\
n^* &= \arg\max_{n\in\{1,\ldots,N\}} J^{(n)}.
\end{aligned}
\end{equation}

Through this process, A$\rightarrow$W conditions future prediction on candidate actions, while W$\rightarrow$A evaluates the predicted outcomes to guide action selection. During closed-loop execution, only the first segment of the selected action sequence is executed, and planning is repeated using new observations.

% 训练目标
\subsection{Learning Objectives}
% 为学习上述共享与私有因素，并实现两种生成机制的有效建模，CMC-WAM 采用由生成任务监督与因素分解约束组成的联合优化目标。在后训练阶段，预训练 WAM 的原有参数保持冻结，优化过程主要作用于新增的表示投影、机制路由器及专家网络。首先，为保持与基础 WAM 的生成范式兼容，我们沿用其原有的动作生成与未来世界预测目标，分别记为 (\mathcal{L}_A) 和 (\mathcal{L}_W)。由此得到基础生成任务损失：

% \mathcal{L}_A+\lambda_W\mathcal{L}_W,
% ]

% 其中，(\lambda_W) 用于平衡动作生成与世界预测任务。两项任务损失分别监督动作与世界分支，同时通过反向传播优化对应的机制路由器和专家网络，使其学习适应各自生成任务的机制选择与映射关系。

% 其次，为约束共享与私有交互因素的组织，引入跨分支一致性损失与冗余抑制损失。前者促进两个分支提取的共同信息保持一致：

% \frac{1}{2}d(q_W(C_t^W),\operatorname{sg}(C_t^A))
% +
% \frac{1}{2}d(q_A(C_t^A),\operatorname{sg}(C_t^W)),
% ]

% 其中，(C_t^A) 和 (C_t^W) 分别表示动作与世界分支提取的共享表示，(q_A) 和 (q_W) 为对应的投影函数，(\operatorname{sg}(\cdot)) 表示停止梯度操作，(d(\cdot,\cdot)) 为表示距离函数。后者通过降低共享交互因素与各分支私有因素之间的线性相关性，减少表示冗余：

% |\operatorname{Cov}(C_t,P_t^A)|_F^2
% +
% |\operatorname{Cov}(C_t,P_t^W)|_F^2,
% ]

% 其中，(\operatorname{Cov}(\cdot,\cdot)) 表示基于训练批次计算的协方差矩阵。结合共享表示的容量约束，上述目标促进共同交互信息与角色特有信息的合理分配。私有因素所需的信息则由各自的生成任务监督予以保留。

% 综合上述目标，CMC-WAM 的整体优化目标定义为：

% [
% \boxed{
% \begin{aligned}
% \mathcal{L}_{\mathrm{total}}
% =&\mathcal{L}A+\lambda_W\mathcal{L}W\
% &+\lambda_C\left(
% \mathcal{L}{\mathrm{share}}
% +\eta\mathcal{L}{\mathrm{red}}
% \right),
% \end{aligned}
% }
% ]

% 其中，(\lambda_C) 和 (\eta) 用于平衡因素分解约束与生成任务监督。通过该目标，CMC-WAM 在保持基础 WAM 生成能力的同时，学习具有共享信息和角色特有信息的结构化表示，并优化两种独立的机制路由与生成网络。

To learn the aforementioned shared and private factors and effectively model the two generative mechanisms, CMC-WAM adopts a joint optimization objective consisting of generative task supervision and factorization constraints. During post-training, the parameters of the pretrained WAM are kept frozen, while optimization is applied primarily to the newly introduced representation projections, mechanism routers, and expert networks.

First, to maintain compatibility with the generative paradigm of the base WAM, we retain its original action generation and future world prediction objectives, denoted by $\mathcal{L}_A$ and $\mathcal{L}_W$, respectively. The resulting generative task objective is:
\begin{equation}
\mathcal{L}_{\mathrm{task}}
=
\mathcal{L}_A+\lambda_W\mathcal{L}_W,
\end{equation}
where $\lambda_W$ balances the action generation and world prediction objectives. These two losses supervise their respective branches and jointly optimize the corresponding mechanism routers and expert networks through backpropagation, enabling them to learn mechanism selection and generative mappings suited to their respective tasks.

Second, to structure the shared and private interaction factors, we introduce cross-branch consistency and redundancy suppression losses. The former encourages the shared representations extracted from the two branches to capture consistent common information:
\begin{equation}
\begin{aligned}
\mathcal{L}_{\mathrm{share}}
=&\frac{1}{2}d(q_W(C_t^W),\operatorname{sg}(C_t^A))\\
&+\frac{1}{2}d(q_A(C_t^A),\operatorname{sg}(C_t^W)),
\end{aligned}
\end{equation}
where $C_t^A$ and $C_t^W$ denote the shared representations extracted from the action and world branches, respectively. $q_A$ and $q_W$ are the corresponding projection functions, $\operatorname{sg}(\cdot)$ denotes stop-gradient, and $d(\cdot,\cdot)$ measures the distance between representations. The latter reduces redundancy by penalizing the linear correlations between the shared interaction state and the private factors of each branch:
\begin{equation}
\mathcal{L}_{\mathrm{red}}
=
\|\operatorname{Cov}(C_t,P_t^A)\|_F^2
+
\|\operatorname{Cov}(C_t,P_t^W)\|_F^2,
\end{equation}
where $\operatorname{Cov}(\cdot,\cdot)$ denotes the covariance matrix computed over a training batch. Together with the capacity constraint on the shared representation, these objectives encourage a structured allocation of common interaction information and role-specific information. The task-relevant information required by the private factors is further preserved through their respective generative task supervision.

Combining these objectives, the overall optimization objective of CMC-WAM is defined as:
\begin{equation}
\boxed{
\begin{aligned}
\mathcal{L}_{\mathrm{total}}
=&\mathcal{L}_A+\lambda_W\mathcal{L}_W\\
&+\lambda_C\left(
\mathcal{L}_{\mathrm{share}}
+\eta\mathcal{L}_{\mathrm{red}}
\right),
\end{aligned}
}
\end{equation}
where $\lambda_C$ and $\eta$ balance the factorization constraints against the generative task objectives. Through this joint optimization, CMC-WAM learns representations containing shared and role-specific information while optimizing the two independent mechanism routing and generation networks, while W$\rightarrow$A guides candidate selection during inference.

\section{Experiments}
This section evaluates CMC-WAM from four aspects:
(1) whether it improves closed-loop manipulation performance
on simulation benchmarks;
(2) how its individual components contribute to the overall
performance;
(3) whether the learned representations and mechanisms
exhibit the intended specialization and coordination;
and (4) how it performs in terms of robustness, computational
efficiency, and real-world applicability.
\begin{table}[t]
    \centering
    
    \caption{
        \textbf{RoboCasa benchmark results.}
        Mean success rates (\%) and standard deviations over three independent evaluation runs across 24 manipulation tasks. All methods are reproduced and evaluated by us under the same RoboCasa evaluation protocol.
    }

    \label{tab:robocasa_main}
    \small
    \setlength{\tabcolsep}{6pt}
    \renewcommand{\arraystretch}{1.08}
    \begin{tabular*}{\columnwidth}{@{\extracolsep{\fill}}lcc@{}}
        \toprule
        Method & Demos/Task & SR (\%) $\uparrow$ \\
        \midrule
        UVA~\cite{li2025unifiedvideoactionmodel}                  & 50   & $49.4 \pm 1.1$ \\
        UWM~\cite{zhu2025unified}                  & 1000 & $60.2 \pm 0.4$ \\
        $\pi_0$~\cite{black2026pi0visionlanguageactionflowmodel}              & 300  & $61.8 \pm 0.8$ \\
        GR00T-N1.5~\cite{nvidia2025gr00tn1openfoundation}           & 300  & $63.6 \pm 0.3$ \\
        Video Policy~\cite{liang2025video}         & 300  & $65.4 \pm 0.5$ \\
        FLARE~\cite{zheng2025flarerobotlearningimplicit}                & 300  & $66.3 \pm 0.3$ \\
        \midrule
        Cosmos Policy-Predict2.5\cite{kim2026cosmos} & 50   & $71.5 \pm 0.2$ \\
        \textbf{CMC-WAM (Ours)} & 50 & $\mathbf{74.2 \pm 0.3}$ \\
        \bottomrule
    \end{tabular*}
\end{table}

\subsection{Experimental Settings}
\label{sec:experimental_settings}
\paragraph{Base Model.}
We adopt Cosmos-Predict2.5~\cite{kim2026cosmos} as the base WAM. It jointly models visual observations, robot states, actions, and future-world information within a shared DiT backbone.% , where Action- and World-related representations are processed by the same Transformer blocks. % For each benchmark, we initialize CMC-WAM from the corresponding Cosmos-Predict2.5-2B robot-policy checkpoint. During post-training, we freeze the pretrained DiT backbone and generation heads, optimizing only the newly introduced CMC modules.

% \paragraph{Benchmarks and Evaluation Protocol.}
% We evaluate CMC-WAM on two robotic manipulation benchmarks: RoboCasa~\cite{nasiriany2024robocasa} and LIBERO~\cite{liu2023libero}. RoboCasa covers all 24 manipulation tasks in the official Cosmos Policy evaluation, while LIBERO comprises four suites (Spatial, Object, Goal, and Long) with 10 tasks each. Following the official Cosmos Policy evaluation protocol, we conduct 50 closed-loop episodes per task for each run.

% All methods are evaluated under consistent protocols within each benchmark. At each decision step, the model generates an action sequence conditioned on the current observations, task instruction, and proprioceptive state. The actions are executed in the environment, and the resulting observations are fed back for subsequent predictions. For controlled comparisons, we use matched environment initializations and perform paired evaluations on the common set of valid episodes, reporting the actual number of valid episodes.

% Each model configuration is evaluated over three independent runs with different environment random seeds. This results in 1,200 and 2,000 episodes per run for RoboCasa and LIBERO, respectively. We report the mean and sample standard deviation of success rates across the three runs. For RoboCasa, we average task-level success rates across all 24 tasks within each run. For LIBERO, we first average the success rates over the 10 tasks within each suite, and then compute the overall average across the four suites. The final results are reported as mean $\pm$ standard deviation.
\paragraph{Benchmarks and Evaluation Protocol.}
We evaluate CMC-WAM on two benchmarks: RoboCasa~\cite{nasiriany2024robocasa}, covering 24 tasks, and LIBERO~\cite{liu2023libero}, with four suites of 10 tasks each. Following the Cosmos Policy evaluation setting, we conduct 50 closed-loop episodes per task in each run, feeding the resulting observations back for subsequent action generation. Each configuration is evaluated over three independent runs with different random seeds, yielding 1,200 episodes per run on RoboCasa and 2,000 on LIBERO. We report mean $\pm$ sample standard deviation across runs. For RoboCasa, we average task-level success rates within each run. For LIBERO, we first average the 10 task-level success rates within each suite and then compute the overall average across the four suites. For paired comparisons between CMC-WAM and Cosmos Policy, we use matched environment initializations and paired evaluations on common valid episodes. Other baselines follow their corresponding benchmark-specific evaluation protocols.

\paragraph{Baselines.}
% We use the original Cosmos-Predict2.5 model without additional mechanism modeling as our primary baseline (Cosmos Policy). CMC-WAM is initialized from the same benchmark-specific checkpoint and introduces only the proposed CMC modules. Both models are evaluated using identical task configurations and closed-loop protocols, providing a controlled comparison of the proposed mechanism modeling approach.

% For broader performance comparison, we reproduce representative methods on both benchmarks. On RoboCasa, these include UVA~\cite{li2025unifiedvideoactionmodel}, UWM~\cite{zhu2025unified}, $\pi_0$~\cite{black2026pi0visionlanguageactionflowmodel}, GR00T-N1.5~\cite{nvidia2025gr00tn1openfoundation}, Video Policy~\cite{liang2025video}, and FLARE~\cite{zheng2025flarerobotlearningimplicit}. On LIBERO, we evaluate $\pi_0$, UniVLA, $\pi_{0.5}$, OpenVLA-OFT, and CogVLA using their respective released checkpoints. All methods are evaluated under the corresponding benchmark-specific closed-loop protocols. We report results from three independent evaluation runs for a consistent comparison.
We use the original Cosmos-Predict2.5\cite{kim2026cosmos} without CMC modules as our primary baseline. CMC-WAM uses the same benchmark-specific initialization and closed-loop evaluation protocols. For broader comparison, we evaluate against UVA~\cite{li2025unifiedvideoactionmodel}, UWM~\cite{zhu2025unified}, $\pi_0$~\cite{black2026pi0visionlanguageactionflowmodel}, GR00T-N1.5~\cite{nvidia2025gr00tn1openfoundation}, Video Policy~\cite{liang2025video}, and FLARE~\cite{zheng2025flarerobotlearningimplicit} on RoboCasa, and $\pi_0$, UniVLA~\cite{bu2505learning}, $\pi_{0.5}$~\cite{intelligence2025pi_}, OpenVLA-OFT~\cite{kim2025fine}, and CogVLA~\cite{li2026cogvla} on LIBERO. We evaluate all methods using their available implementations or released checkpoints under the corresponding benchmark protocols. All reported results are obtained from our own evaluations. Further details are provided in Appendix~\ref{app:baselines}.

\begin{table*}[t]
    \centering
    \caption{
        \textbf{LIBERO benchmark results.}
        Mean success rates (\%) and standard deviations over three independent evaluation runs across four manipulation suites.
    }
    \label{tab:libero_main}
    \small
    \setlength{\tabcolsep}{6pt}
    \renewcommand{\arraystretch}{1.08}
    \begin{tabular*}{\textwidth}{@{\extracolsep{\fill}}lccccc@{}}
        \toprule
        Method & Spatial & Object & Goal & Long & Average $\uparrow$ \\
        \midrule
        $\pi_0$~\cite{black2026pi0visionlanguageactionflowmodel}
        & $96.4 \pm 0.5$ & $98.2 \pm 0.8$ & $95.2 \pm 1.1$ & $84.6 \pm 1.8$ & $93.60 \pm 0.56$ \\
        UniVLA~\cite{bu2505learning}
        & $96.2 \pm 0.6$ & $96.5 \pm 1.0$ & $94.9 \pm 0.8$ & $91.8 \pm 1.3$ & $94.85 \pm 0.50$ \\
        $\pi_{0.5}$~\cite{intelligence2025pi_}
        & $98.2 \pm 0.6$ & $97.9 \pm 0.5$ & $98.1 \pm 0.7$ & $92.2 \pm 1.0$ & $96.60 \pm 0.38$ \\
        OpenVLA-OFT~\cite{kim2025fine}
        & $97.3 \pm 0.5$ & $98.1 \pm 0.4$ & $97.2 \pm 0.3$ & $93.5 \pm 1.1$ & $96.53 \pm 0.35$ \\
        CogVLA~\cite{li2026cogvla}
        & $98.4 \pm 0.7$ & $98.2 \pm 0.6$ & $96.3 \pm 0.9$ & $95.1 \pm 1.1$ & $97.00 \pm 0.44$ \\
        \midrule
        Cosmos Policy-Predict2.5~\cite{kim2026cosmos}
        & $98.5 \pm 0.4$ & $99.1 \pm 0.2$ & $98.0 \pm 0.4$ & $96.2 \pm 1.0$ & $97.95 \pm 0.28$ \\
        \textbf{CMC-WAM (Ours)}
        & \bm{$98.6 \pm 0.4$}
        & \bm{$99.3 \pm 0.5$}
        & \bm{$98.2 \pm 0.6$}
        & \bm{$97.2 \pm 0.9$}
        & \bm{$98.33 \pm 0.30$} \\
        \bottomrule
    \end{tabular*}
\end{table*}

\paragraph{Implementation Details.}
% We implement CMC-WAM on Cosmos-Predict2.5-2B, extracting action and world representations from slot 5 and slots 6-9, respectively, at the output of DiT block indexed 13. Each branch employs eight two-layer MLP experts with independent Top-2 routing. The experts predict residual corrections to the pretrained flow-matching velocity fields, with zero-initialized residual outputs. We freeze the pretrained DiT backbone and generation heads, optimizing only the CMC modules.

% We use the RoboCasa demonstration dataset and train with AdamW, a learning rate of $10^{-3}$, and an effective batch size of 4 FM training windows. During inference, we sample four candidate action sequences of length 32 and predict their corresponding future world states at $t+32$. Candidates are evaluated using the pretrained Cosmos value function, and the first 16 steps of the highest-valued action sequence are executed before replanning.
We implement CMC-WAM on Cosmos-Predict2.5-2B, extracting the action representation from slot 5 and world representations from slots 6--9 at the output of DiT block 13(zero-indexed). Each branch employs eight two-layer MLP experts with independent Top-2 routing, whose zero-initialized outputs provide residual corrections to the pretrained flow-matching velocity fields. The pretrained DiT backbone and generation heads remain frozen, with only the CMC modules optimized. During inference, we generate four candidate action sequences of length 32, predict their corresponding future states, and evaluate them using the pretrained Cosmos value function. The first 16 steps of the highest-valued candidate are executed before replanning. Further implementation and training details are provided in Appendix~\ref{app:implementation}.

\subsection{Main Results}

\paragraph{Results on RoboCasa.}
As shown in Table~\ref{tab:robocasa_main}, CMC-WAM achieves an average success rate of 74.2\% on RoboCasa, outperforming the original Cosmos Policy (71.5\%) by 2.7 percentage points. This corresponds to approximately 32 additional successful episodes per 1,200-episode evaluation run. CMC-WAM also surpasses the strongest reproduced non-Cosmos baseline, FLARE (66.3\%), by 7.9 percentage points, demonstrating its effectiveness in improving closed-loop manipulation performance.

\paragraph{Results on LIBERO.}
As shown in Table~\ref{tab:libero_main}, CMC-WAM achieves an average success rate of 98.33\% across the four LIBERO suites, improving upon Cosmos Policy (97.95\%) by 0.38 percentage points. The largest gain is observed on the Long suite, with a 1.0 percentage-point improvement. This suggests that CMC-WAM is particularly beneficial for tasks involving longer interaction sequences.

Additional task-wise results on RoboCasa and LIBERO are provided in Appendix~\ref{app:Results}.

\subsection{Ablation Studies}

We conduct ablation studies on RoboCasa to examine the contributions of the key components in CMC-WAM. As shown in Table~\ref{tab:ablation}, adding W$\rightarrow$A prediction-guided selection alone yields a marginal gain of 0.1 percentage points over Cosmos Policy. In contrast, Full CMC-WAM achieves a further improvement of 2.4 percentage points over this variant. Removing factorization, replacing independent routers with a shared router, and removing A$\rightarrow$W or W$\rightarrow$A coupling lead to performance drops of 1.8, 1.2, 1.3, and 1.9 percentage points, respectively. These results highlight the complementary contributions of shared/private factorization, independent mechanism routing, and asymmetric coupling. Further details of the ablation configurations and evaluation protocol are provided in Appendix~\ref{app:Ablation}.

\begin{table}[t]
    \centering
    \caption{
        \textbf{Ablation studies of CMC-WAM on RoboCasa.}
        Success rates (\%) from a single evaluation run
        under matched environment initializations.
    }
    \label{tab:ablation}
    \small
    \setlength{\tabcolsep}{5pt}
    \renewcommand{\arraystretch}{1.12}
    \begin{tabular*}{\columnwidth}{@{\extracolsep{\fill}}lc@{}}
        \toprule
        Configuration & SR (\%) $\uparrow$ \\
        \midrule
        Original Cosmos Policy
            & 71.5 \\
        Cosmos + W$\rightarrow$A (no CMC)
            & 71.6 \\
        \midrule
        w/o Factorization
            & 72.2 \\
        Shared Router
            & 72.8 \\
        \midrule
        w/o A$\rightarrow$W Coupling
            & 72.7 \\
        w/o W$\rightarrow$A Coupling
            & 72.1 \\
        \midrule
        \textbf{Full CMC-WAM}
            & \textbf{74.0} \\
        \bottomrule
    \end{tabular*}
\end{table}

\subsection{Mechanism Analysis}
% 解释 CMC-WAM 为什么有效：
% 1. 分析共享/私有因素是否表现出不同的信息角色；
% 2. 分析 Action/World Router 在不同交互阶段的动态响应；
% 3. 验证不同动作条件下的未来世界预测能力。
% 通过一张综合可视化图和必要的定量指标提供机制证据。
% \label{sec:mechanism_analysis}

%To further understand the effectiveness of CMC-WAM, we analyze factor specialization, dynamic routing, and mechanism coupling. The shared factor achieves an $R^2$ of 0.70 for predicting common interaction states. Relative to the shared factor, the Action and World Private factors achieve $R^2$ gains of 0.13 and 0.16 on action prediction and future-state prediction, respectively. Both routers exhibit stage-dependent changes in their routing distributions, with average pairwise Jensen--Shannon divergences of 0.25 and 0.32 for the Action and World routers, respectively. Regarding mechanism coupling, matched action conditions reduce future-state prediction error by 0.049 compared with shuffled conditions. The predicted future states also achieve a Pairwise Ranking Accuracy of 78.8\% in candidate-action comparisons. Together, these results provide complementary evidence for factor specialization, adaptive mechanism routing, and asymmetric coupling. Detailed analyses are provided in Appendix~\ref{app:mechanism_analysis}.
To further understand the effectiveness of CMC-WAM, we analyze factor specialization, dynamic routing, and mechanism coupling. The shared factor achieves an $R^2$ of 0.70 for predicting common interaction states. Compared with the shared factor, the Action and World Private factors achieve $R^2$ gains of 0.13 and 0.16 on action prediction and future-state prediction, respectively. Both routers exhibit stage-dependent changes in their routing distributions, with average pairwise Jensen--Shannon divergences of 0.25 and 0.32 for the Action and World routers, respectively. Regarding mechanism coupling, matched action conditions reduce future-state prediction error by 0.049 compared with shuffled conditions. The predicted future states also achieve a Pairwise Ranking Accuracy of 78.8\% in candidate-action comparisons. Together, these results provide complementary evidence for factor specialization, adaptive mechanism routing, and asymmetric coupling. Detailed analyses are provided in Appendix~\ref{app:mechanism_analysis}.

\subsection{Generalization and Efficiency}

To assess robustness beyond the standard setting, we evaluate CMC-WAM under shifted initial configurations. It achieves a success rate of 66.7\%, compared with 61.7\% for Cosmos Policy, widening the performance gap from 2.7 to 5.0 percentage points. In terms of computational cost, CMC-WAM introduces only 0.35M additional trainable parameters (0.018\% of the 2B backbone), while four-candidate W$\rightarrow$A planning adds 390 ms per decision (+172\%). The complete generalization and efficiency results are reported in Appendix~\ref{app:Generalization}.

\subsection{Real-World Experiments}

We further evaluate CMC-WAM on a physical dual-arm platform across three manipulation tasks. Over 90 independent trials, it achieves an overall success rate of 84.4\%, with 86.7\% for cup handover, 76.7\% for plate insertion, and 90.0\% for remote control placement. These results support the applicability of CMC-WAM to real-world robotic manipulation. Further details of the platform and evaluation protocol are provided in Appendix~\ref{app:real_world}.

\section{Conclusion}

We present CMC-WAM, which investigates and models the structural dependencies between action generation and world evolution in WAMs from a causal mechanism perspective. Through shared and private factorization, independent mechanism routing, and asymmetric mechanism coupling, CMC-WAM enables structured modeling and coordination of the two generative processes within a unified architecture. Closed-loop experiments on RoboCasa and LIBERO achieve higher mean success rates and enhanced robustness under shifted initial configurations. Real-world experiments further validate its applicability to diverse physical manipulation tasks. Future work will focus on reducing the computational overhead of online prediction and planning, and exploring generalization to more complex tasks and robotic platforms.

\section*{LLM Usage Statement}
We used OpenAI's ChatGPT to provide reviewer-style feedback on the manuscript and to assist in refining author-produced content related to writing, related work and citations, code and implementation, and experiments. No LLM tools were used to generate research ideas, hypotheses, scientific claims, conclusions, figures, or datasets.

{
    \small
    \bibliographystyle{ieeenat_fullname}
    \bibliography{Chapters/6_references}
}
\newpage
% \section*{Appendix}
\setcounter{section}{0}
\renewcommand{\thesection}{\Alph{section}}
\clearpage
\setcounter{page}{1}
\section*{Appendix}
% 附录
The appendix is organized as follows:
\begin{enumerate}
    \item \textbf{Appendix~\ref{app:implementation}} provides implementation details of CMC-WAM, including its architecture, feature extraction, training configuration, and closed-loop inference.

    \item \textbf{Appendix~\ref{app:Results}} presents additional benchmark results, including task-wise performance on RoboCasa and LIBERO and results across independent evaluation runs.
    
    \item \textbf{Appendix~\ref{app:baselines}} briefly describes the baseline methods included in our comparisons.
    
    \item \textbf{Appendix~\ref{app:Ablation}} provides detailed ablation configurations and supplementary results to further examine the contributions of individual components.

    \item \textbf{Appendix~\ref{app:mechanism_analysis}} presents detailed mechanism analyses of factor specialization, dynamic routing, and action-conditioned world prediction.

    \item \textbf{Appendix~\ref{app:Generalization}} provides additional analyses of generalization under distribution shifts, candidate selection, and computational efficiency.

    \item \textbf{Appendix~\ref{app:real_world}} describes the real-world robotic platform, manipulation tasks, and experimental protocols.

    \item \textbf{Appendix~\ref{app:limitations}} discusses the limitations of CMC-WAM and directions for future research.
\end{enumerate}

\section{Implementation Details}
\label{app:implementation}

\subsection{CMC Architecture and Feature Extraction}

\paragraph{Modality Encoding.}
CMC-WAM follows the pretrained modality-encoding pipeline of Cosmos-Predict2.5 without introducing additional modality encoders. Language instructions are encoded by Cosmos-Reason1 and provided to the DiT through cross-attention. Visual observations are encoded into spatiotemporal latents using the causal Wan2.1 VAE, with 4$\times$ temporal and 8$\times$ spatial compression. The resulting latents are further processed by $1\times2\times2$ patchification. Robot actions and proprioceptive states are normalized to $[-1,1]$ and injected into designated latent frames through deterministic repetition-based mapping.

\paragraph{Feature Extraction.}
We extract unified hidden representations from the output of DiT block 13 (zero-indexed) in Cosmos-Predict2.5-2B. Specifically, slot 5 provides the Action features, while slots 6--9 provide the World features. We apply mean pooling to aggregate the corresponding tokens:

\begin{equation}
\begin{aligned}
H_t^A &= \operatorname{MeanPool}_{\mathrm{spatial}}\left(H_{13}^{(5)}\right),\\
H_t^W &= \operatorname{MeanPool}_{\mathrm{slots,spatial}}\left(H_{13}^{(6:9)}\right).
\end{aligned}
\end{equation}

The Action features are averaged over spatial tokens, while the World features are averaged over the four selected slots and their spatial tokens. Both representations are extracted from the same pretrained DiT backbone without separate feature encoders.

\paragraph{Factorization and Mechanism Experts.}
The extracted Action and World representations are processed by role-specific projection layers to obtain 32-dimensional shared and private factors. We combine the shared factors through normalized fusion:
\begin{equation}
C_t = \operatorname{LN}\left(\frac{\operatorname{LN}(C_t^A)+\operatorname{LN}(C_t^W)}{2}\right),
\end{equation}
where $\operatorname{LN}$ denotes layer normalization without learnable affine parameters. This operation normalizes the shared factors before fusion to reduce scale differences between the two branches.

The Action and World branches employ independent mechanism routers, each with eight two-layer MLP experts. Both routers use Top-2 sparse routing, with routing weights renormalized over the selected experts. The weighted expert outputs provide residual corrections to the pretrained flow-matching velocity fields. The residual output layers are zero-initialized, preserving the original velocity predictions at initialization.

The CMC modules introduce approximately 0.35M trainable parameters, accounting for about 0.018\% of the 2B backbone parameters. The pretrained DiT backbone and generation heads remain frozen throughout training.

\subsection{Training Configuration}

For each benchmark, we follow the corresponding Cosmos-Predict2.5 training-data configuration and demonstration sample budget. We optimize the CMC-specific projection layers, routers, and experts using AdamW with an initial learning rate of $10^{-3}$ and weight decay of $10^{-4}$. We adopt a cosine learning rate scheduler and train for 800 steps. The micro-batch size is four flow-matching training windows, with gradient accumulation set to one, resulting in an effective batch size of four.

We optimize the joint objective defined in Eq.~(14), combining the original action generation and world prediction losses with the cross-branch consistency and redundancy suppression constraints. We set $\lambda_W=1.0$, $\lambda_C=0.1$, and $\eta=10^{-3}$. The distance function in the shared consistency loss is defined as the mean squared error over feature dimensions. Only the CMC projection layers, routers, and experts are updated, while the pretrained backbone and generation heads remain frozen.

\subsection{Closed-loop Inference}

At each decision step, CMC-WAM generates four candidate action sequences from independently sampled initial noise, each with a horizon of 32 steps. For each candidate, the World branch predicts the corresponding future state over 32 steps, conditioned on the candidate action sequence. The pretrained Cosmos Value Function evaluates the predicted terminal state, and the candidate with the highest predicted value is selected for execution.

Only the first 16 actions of the selected sequence are executed before replanning. The robot then receives updated observations and repeats candidate generation, future prediction, and value-based selection. This World-to-Action (W$\rightarrow$A) mechanism is enabled by default during CMC-WAM inference and is disabled only for the corresponding ablation studies.

\section{Benchmark Evaluation and Additional Results}
\label{app:Results}

\subsection{Evaluation Protocol}
\label{app:eval_protocol}

\paragraph{Benchmark Settings.}
We evaluate CMC-WAM and the baselines on two benchmarks: RoboCasa, which covers 24 manipulation tasks, and LIBERO, which consists of four suites (Spatial, Object, Goal, and Long), each containing 10 tasks. For each method, we conduct 50 valid closed-loop evaluation episodes per task in each independent run. During evaluation, the policy receives updated observations after each action execution for subsequent decision-making. Each configuration is evaluated with three independent random seeds, resulting in 1,200 and 2,000 valid episodes per run on RoboCasa and LIBERO, respectively.

\paragraph{Success Rate.}
We use the task success rate (SR) as the primary evaluation metric, calculated as the proportion of successful episodes among all valid evaluation episodes. For RoboCasa, we first calculate the success rate of each task and then take the unweighted average across the 24 tasks within each run. For LIBERO, we first average the success rates across the 10 tasks within each suite, followed by an unweighted average across the four suites. We report the mean and sample standard deviation (Mean $\pm$ Sample SD) across the three independent runs to characterize performance and variability across random seeds.

\paragraph{Episode Validity and Paired Evaluation.}
Episodes affected by external evaluation failures, such as simulator crashes, rendering errors, or corrupted observations, are excluded and replaced with new episodes until the target number of valid episodes is reached. This protocol is consistently applied across all methods and benchmarks. In contrast, policy-related execution failures, invalid actions, and unsuccessful task completions are retained and counted as failures. For direct comparisons between CMC-WAM and Cosmos Policy, both methods are evaluated using matched environment initializations and paired evaluations on common valid episodes. If an external evaluation failure invalidates an initialization, it is excluded from both methods and replaced with a new shared initialization. Other baselines follow their respective benchmark-specific evaluation protocols.
% Appendix B. Additional Benchmark Results

% B.1 Task-wise RoboCasa Results
% - 24 个任务分别的 SR；
% - Cosmos Policy 与 CMC-WAM 的逐任务比较；
% - 标记性能提升和退化的任务。
\subsection{Task-wise RoboCasa Results}
\label{app:robocasa_results}
% B.2 Task-wise LIBERO Results
% - Spatial、Object、Goal、Long 四个 Suite 的逐任务 SR；
% - 40 个任务的详细对比；
% - 重点展示 Long Suite 的性能变化。
To provide a more detailed comparison, we report the task-wise success rates of CMC-WAM and the original Cosmos-Predict2.5 across all 24 RoboCasa tasks in Table~\ref{tab:robocasa_taskwise}. Each task consists of 50 valid evaluation episodes, yielding 1,200 episodes per method. Overall, CMC-WAM achieves a success rate of 74.00\% (888/1,200), compared with 71.50\% (858/1,200) for the original Cosmos-Predict2.5, representing an improvement of 2.50 percentage points. At the task level, CMC-WAM improves performance on 22 out of 24 tasks, while maintaining the same success rate on the remaining two tasks. These results provide a detailed view of the performance gains across individual manipulation tasks. The main-text results are based on three independent runs and are reported as mean $\pm$ sample standard deviation.

\begin{table*}[t]
\centering
\caption{Task-wise success rates (\%) on RoboCasa. Each task is evaluated over 50 valid episodes in one evaluation run. $\Delta$SR denotes the improvement of CMC-WAM over the original Cosmos-Predict2.5 in percentage points.}
\label{tab:robocasa_taskwise}
\small
\setlength{\tabcolsep}{8pt}
\renewcommand{\arraystretch}{1.12}
\begin{tabular}{r l c c c}
\hline
\textbf{\#} &
\textbf{Task} &
\textbf{Cosmos-Predict2.5} &
\textbf{CMC-WAM} &
\textbf{$\Delta$SR (pp)} \\
\hline
1  & PnPCounterToCab        & 30/50 (60.0\%)  & 32/50 (64.0\%)  & +4.0 \\
2  & PnPCabToCounter        & 22/50 (44.0\%)  & 24/50 (48.0\%)  & +4.0 \\
3  & PnPCounterToSink       & 29/50 (58.0\%)  & 30/50 (60.0\%)  & +2.0 \\
4  & PnPSinkToCounter       & 35/50 (70.0\%)  & 36/50 (72.0\%)  & +2.0 \\
5  & PnPCounterToMicrowave  & 18/50 (36.0\%)  & 20/50 (40.0\%)  & +4.0 \\
6  & PnPMicrowaveToCounter  & 24/50 (48.0\%)  & 26/50 (52.0\%)  & +4.0 \\
7  & PnPCounterToStove      & 36/50 (72.0\%)  & 37/50 (74.0\%)  & +2.0 \\
8  & PnPStoveToCounter      & 29/50 (58.0\%)  & 30/50 (60.0\%)  & +2.0 \\
9  & OpenSingleDoor         & 47/50 (94.0\%)  & 48/50 (96.0\%)  & +2.0 \\
10 & CloseSingleDoor        & 46/50 (92.0\%)  & 47/50 (94.0\%)  & +2.0 \\
11 & OpenDoubleDoor         & 49/50 (98.0\%)  & 50/50 (100.0\%) & +2.0 \\
12 & CloseDoubleDoor        & 37/50 (74.0\%)  & 38/50 (76.0\%)  & +2.0 \\
13 & OpenDrawer             & 46/50 (92.0\%)  & 47/50 (94.0\%)  & +2.0 \\
14 & CloseDrawer            & 50/50 (100.0\%) & 50/50 (100.0\%) & 0.0 \\
15 & TurnOnStove            & 32/50 (64.0\%)  & 34/50 (68.0\%)  & +4.0 \\
16 & TurnOffStove           & 17/50 (34.0\%)  & 19/50 (38.0\%)  & +4.0 \\
17 & TurnOnSinkFaucet       & 39/50 (78.0\%)  & 40/50 (80.0\%)  & +2.0 \\
18 & TurnOffSinkFaucet      & 44/50 (88.0\%)  & 45/50 (90.0\%)  & +2.0 \\
19 & TurnSinkSpout          & 43/50 (86.0\%)  & 44/50 (88.0\%)  & +2.0 \\
20 & CoffeeSetupMug         & 17/50 (34.0\%)  & 19/50 (38.0\%)  & +4.0 \\
21 & CoffeeServeMug         & 31/50 (62.0\%)  & 33/50 (66.0\%)  & +4.0 \\
22 & CoffeePressButton      & 48/50 (96.0\%)  & 49/50 (98.0\%)  & +2.0 \\
23 & TurnOnMicrowave        & 39/50 (78.0\%)  & 40/50 (80.0\%)  & +2.0 \\
24 & TurnOffMicrowave       & 50/50 (100.0\%) & 50/50 (100.0\%) & 0.0 \\
\hline
\textbf{Average} & &
\textbf{858/1200 (71.50\%)} &
\textbf{888/1200 (74.00\%)} &
\textbf{+2.50} \\
\hline
\end{tabular}
\end{table*}

\subsection{Task-wise LIBERO Results}
\label{app:libero_results}
% B.2 Results across Independent Runs
% - 三个 Seed 的原始 SR；
% - 每个 Suite 或 Benchmark 的标准差计算结果；
% - 有效 Episode 数量及成功次数。

To further examine the performance of CMC-WAM across different types of manipulation tasks, we report the task-wise results on all 40 tasks in the four LIBERO suites. Table~\ref{tab:libero_taskwise} presents the success counts and success rates of the original Cosmos-Predict2.5 and CMC-WAM on LIBERO-Spatial, LIBERO-Object, LIBERO-Goal, and LIBERO-Long. Each task is evaluated over 50 valid episodes, resulting in 2,000 episodes per method.

Overall, CMC-WAM achieves a success rate of 98.5\% (1970/2000), compared with 97.8\% (1956/2000) for the original Cosmos-Predict2.5, yielding an improvement of 0.7 percentage points. Across the four suites, the improvements are 0.6, 0.2, 0.8, and 1.2 percentage points on Spatial, Object, Goal, and Long, respectively. The largest gain is observed on LIBERO-Long. At the task level, CMC-WAM improves performance on 13 out of 40 tasks and maintains the same success rate on the remaining 27 tasks, with no task-level regressions. The task-wise results in Table~\ref{tab:libero_taskwise} are obtained from a single evaluation run, whereas the main-text results are averaged over three independent runs.

\begin{table*}[t]
\centering
\caption{Task-wise success rates (\%) on LIBERO. Each task is evaluated over 50 valid episodes in one evaluation run. $\Delta$SR denotes the improvement of CMC-WAM over the original Cosmos-Predict2.5 in percentage points.}
\label{tab:libero_taskwise}
\footnotesize
\setlength{\tabcolsep}{4pt}
\renewcommand{\arraystretch}{1.02}
\begin{tabularx}{\textwidth}{@{}c X c c c@{}}
\hline
\textbf{ID} &
\textbf{Task Instruction} &
\textbf{\shortstack{Cosmos\\Predict2.5}} &
\textbf{CMC-WAM} &
\textbf{$\Delta$SR (pp)} \\
\hline

% ==================== LIBERO-Spatial ====================
\multicolumn{5}{l}{\textbf{LIBERO-Spatial}} \\
\hline
S0 & Pick up the black bowl between the plate and the ramekin and place it on the plate
& 49/50 (98.0\%) & 49/50 (98.0\%) & 0.0 \\

S1 & Pick up the black bowl next to the ramekin and place it on the plate
& 50/50 (100.0\%) & 50/50 (100.0\%) & 0.0 \\

S2 & Pick up the black bowl from table center and place it on the plate
& 49/50 (98.0\%) & 49/50 (98.0\%) & 0.0 \\

S3 & Pick up the black bowl on the cookie box and place it on the plate
& 49/50 (98.0\%) & 49/50 (98.0\%) & 0.0 \\

S4 & Pick up the black bowl in the top drawer of the wooden cabinet and place it on the plate
& 49/50 (98.0\%) & 50/50 (100.0\%) & +2.0 \\

S5 & Pick up the black bowl on the ramekin and place it on the plate
& 47/50 (94.0\%) & 49/50 (98.0\%) & +4.0 \\

S6 & Pick up the black bowl next to the cookie box and place it on the plate
& 50/50 (100.0\%) & 50/50 (100.0\%) & 0.0 \\

S7 & Pick up the black bowl on the stove and place it on the plate
& 49/50 (98.0\%) & 49/50 (98.0\%) & 0.0 \\

S8 & Pick up the black bowl next to the plate and place it on the plate
& 49/50 (98.0\%) & 49/50 (98.0\%) & 0.0 \\

S9 & Pick up the black bowl on the wooden cabinet and place it on the plate
& 49/50 (98.0\%) & 49/50 (98.0\%) & 0.0 \\

\multicolumn{2}{r}{\textbf{Spatial Average}} &
\textbf{490/500 (98.0\%)} &
\textbf{493/500 (98.6\%)} &
\textbf{+0.6} \\
\hline

% ==================== LIBERO-Object ====================
\multicolumn{5}{l}{\textbf{LIBERO-Object}} \\
\hline
O0 & Pick up the alphabet soup and place it in the basket
& 49/50 (98.0\%) & 50/50 (100.0\%) & +2.0 \\

O1 & Pick up the cream cheese and place it in the basket
& 50/50 (100.0\%) & 50/50 (100.0\%) & 0.0 \\

O2 & Pick up the salad dressing and place it in the basket
& 50/50 (100.0\%) & 50/50 (100.0\%) & 0.0 \\

O3 & Pick up the BBQ sauce and place it in the basket
& 50/50 (100.0\%) & 50/50 (100.0\%) & 0.0 \\

O4 & Pick up the ketchup and place it in the basket
& 49/50 (98.0\%) & 49/50 (98.0\%) & 0.0 \\

O5 & Pick up the tomato sauce and place it in the basket
& 50/50 (100.0\%) & 50/50 (100.0\%) & 0.0 \\

O6 & Pick up the butter and place it in the basket
& 49/50 (98.0\%) & 49/50 (98.0\%) & 0.0 \\

O7 & Pick up the milk and place it in the basket
& 49/50 (98.0\%) & 49/50 (98.0\%) & 0.0 \\

O8 & Pick up the chocolate pudding and place it in the basket
& 50/50 (100.0\%) & 50/50 (100.0\%) & 0.0 \\

O9 & Pick up the orange juice and place it in the basket
& 49/50 (98.0\%) & 49/50 (98.0\%) & 0.0 \\

\multicolumn{2}{r}{\textbf{Object Average}} &
\textbf{495/500 (99.0\%)} &
\textbf{496/500 (99.2\%)} &
\textbf{+0.2} \\
\hline

% ==================== LIBERO-Goal ====================
\multicolumn{5}{l}{\textbf{LIBERO-Goal}} \\
\hline
G0 & Open the middle drawer of the cabinet
& 48/50 (96.0\%) & 49/50 (98.0\%) & +2.0 \\

G1 & Put the bowl on the stove
& 50/50 (100.0\%) & 50/50 (100.0\%) & 0.0 \\

G2 & Put the wine bottle on top of the cabinet
& 50/50 (100.0\%) & 50/50 (100.0\%) & 0.0 \\

G3 & Open the top drawer and put the bowl inside
& 50/50 (100.0\%) & 50/50 (100.0\%) & 0.0 \\

G4 & Put the bowl on top of the cabinet
& 48/50 (96.0\%) & 49/50 (98.0\%) & +2.0 \\

G5 & Push the plate to the front of the stove
& 50/50 (100.0\%) & 50/50 (100.0\%) & 0.0 \\

G6 & Put the cream cheese in the bowl
& 49/50 (98.0\%) & 50/50 (100.0\%) & +2.0 \\

G7 & Turn on the stove
& 50/50 (100.0\%) & 50/50 (100.0\%) & 0.0 \\

G8 & Put the bowl on the plate
& 50/50 (100.0\%) & 50/50 (100.0\%) & 0.0 \\

G9 & Put the wine bottle on the rack
& 45/50 (90.0\%) & 46/50 (92.0\%) & +2.0 \\

\multicolumn{2}{r}{\textbf{Goal Average}} &
\textbf{490/500 (98.0\%)} &
\textbf{494/500 (98.8\%)} &
\textbf{+0.8} \\
\hline

% ==================== LIBERO-Long ====================
\multicolumn{5}{l}{\textbf{LIBERO-Long}} \\
\hline
L0 & Put both the alphabet soup and the tomato sauce in the basket
& 47/50 (94.0\%) & 48/50 (96.0\%) & +2.0 \\

L1 & Put both the cream cheese box and the butter in the basket
& 48/50 (96.0\%) & 49/50 (98.0\%) & +2.0 \\

L2 & Turn on the stove and put the moka pot on it
& 49/50 (98.0\%) & 49/50 (98.0\%) & 0.0 \\

L3 & Put the black bowl in the bottom drawer of the cabinet and close it
& 49/50 (98.0\%) & 49/50 (98.0\%) & 0.0 \\

L4 & Put the white mug on the left plate and put the yellow and white mug on the right plate
& 48/50 (96.0\%) & 49/50 (98.0\%) & +2.0 \\

L5 & Pick up the book and place it in the back compartment of the caddy
& 49/50 (98.0\%) & 49/50 (98.0\%) & 0.0 \\

L6 & Put the white mug on the plate and put the chocolate pudding to the right of the plate
& 46/50 (92.0\%) & 47/50 (94.0\%) & +2.0 \\

L7 & Put both the alphabet soup and the cream cheese box in the basket
& 48/50 (96.0\%) & 49/50 (98.0\%) & +2.0 \\

L8 & Put both moka pots on the stove
& 49/50 (98.0\%) & 49/50 (98.0\%) & 0.0 \\

L9 & Put the yellow and white mug in the microwave and close it
& 48/50 (96.0\%) & 49/50 (98.0\%) & +2.0 \\

\multicolumn{2}{r}{\textbf{Long Average}} &
\textbf{481/500 (96.2\%)} &
\textbf{487/500 (97.4\%)} &
\textbf{+1.2} \\
\hline

% ==================== Overall ====================
\textbf{Overall Average} & &
\textbf{1956/2000 (97.8\%)} &
\textbf{1970/2000 (98.5\%)} &
\textbf{+0.7} \\
\hline
\end{tabularx}
\end{table*}

\begin{figure*}[t]
    \centering
    \includegraphics[width=\textwidth]{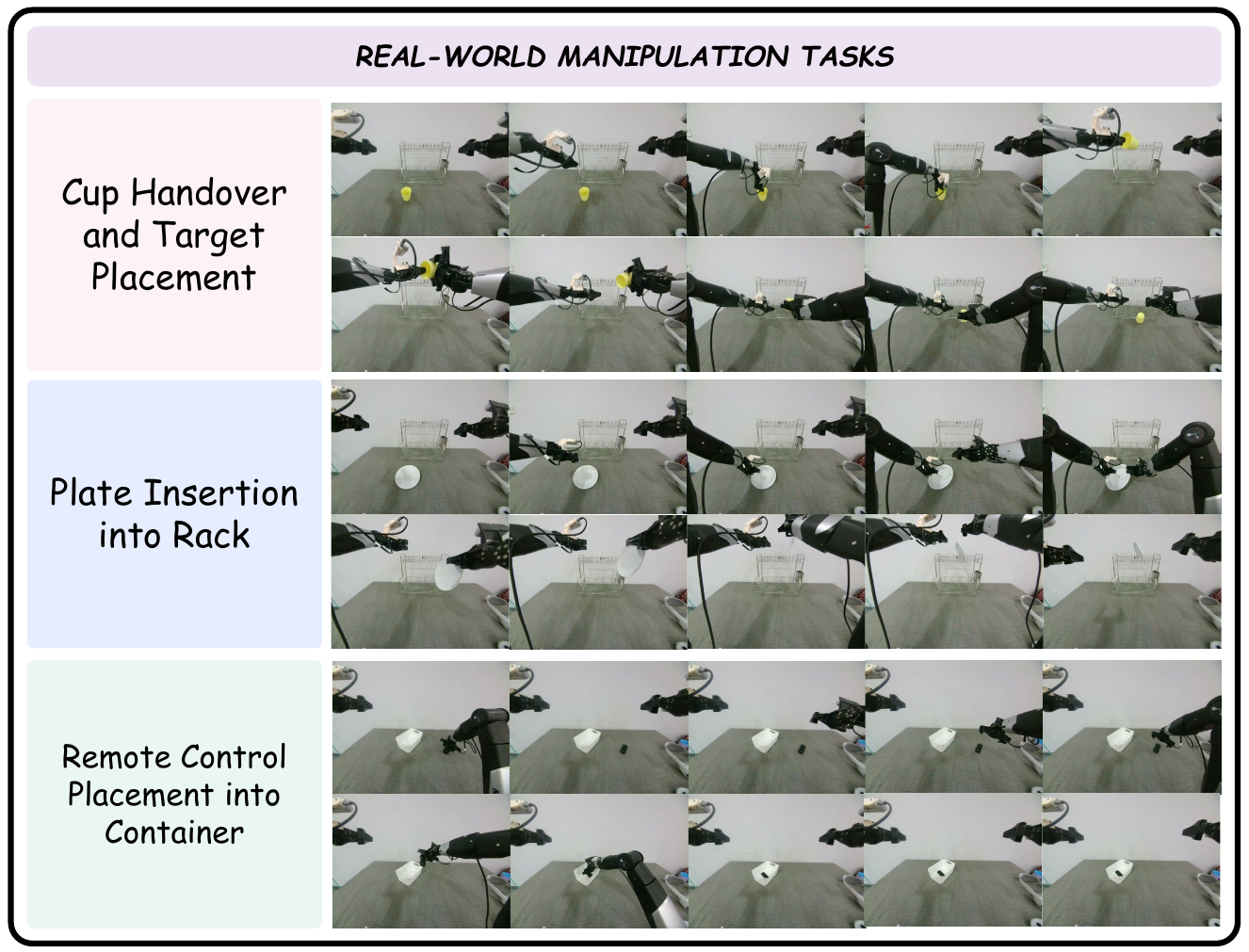}
    \caption{
        \textbf{Real-world robotic manipulation tasks.}
        We evaluate CMC-WAM on three representative tasks:
        (a) cup handover and target placement,
        (b) plate insertion into a rack, and
        (c) remote control placement into a container.
        Each row illustrates the execution sequence of a task.
    }
    \label{fig:real_world_tasks}
\end{figure*}
\section{Baseline Methods}
\label{app:baselines}

We compare CMC-WAM with ten representative baseline methods covering unified video-action modeling, implicit world modeling, and VLA policies. We also include the original Cosmos-Predict2.5 without CMC modules as our primary baseline.

\begin{itemize}

    \item \textbf{Cosmos Policy-Predict2.5 \cite{kim2026cosmos}:}
    A pretrained video model adapted for visuomotor control and planning. It jointly models visual observations, robot states, actions, and future-world information within a shared DiT backbone, serving as our primary baseline.

    \item \textbf{UVA \cite{li2025unifiedvideoactionmodel}:}
    A unified video-action model that jointly learns video prediction and action generation through shared representations. It connects visual dynamics and robot control within a unified generative framework.

    \item \textbf{UWM \cite{zhu2025unified}:}
    A unified world model that couples video and action diffusion within a shared Transformer architecture. It supports joint pretraining on large-scale robotic datasets by modeling visual and action sequences together.

    \item \textbf{Video Policy \cite{liang2025video}:}
    A video-generation-based approach to robot policy learning. It leverages generated visual trajectories to model future robot behavior and support action generation.

    \item \textbf{FLARE \cite{zheng2025flarerobotlearningimplicit}:}
    A robot learning framework with implicit world modeling. It aligns policy features with future visual representations, enabling the policy to anticipate action consequences without explicitly generating future videos.

    \item \textbf{$\pi_0$ \cite{black2026pi0visionlanguageactionflowmodel}:}
    A generalist VLA model based on flow matching. It combines a pretrained vision-language backbone with an action expert to generate continuous robot action sequences.

    \item \textbf{$\pi_{0.5}$ \cite{intelligence2025pi_}:}
    An extension of $\pi_0$ designed for open-world generalization. It incorporates diverse robot and web data to improve task generalization across environments and manipulation scenarios.

    \item \textbf{GR00T-N1.5 \cite{nvidia2025gr00tn1openfoundation}:}
    A generalist humanoid robot foundation model that combines a vision-language backbone with a Diffusion Transformer for action generation, supporting language-conditioned perception and robot control.

    \item \textbf{UniVLA \cite{bu2505learning}:}
    A cross-embodiment VLA model that learns task-centric latent actions from large-scale video data. These latent representations facilitate knowledge transfer across different robot embodiments.

    \item \textbf{OpenVLA-OFT \cite{kim2025fine}:}
    An efficient fine-tuning framework for OpenVLA that incorporates parallel action decoding, action chunking, and continuous action representations to improve inference efficiency and manipulation performance.

    \item \textbf{CogVLA \cite{li2026cogvla}:}
    A cognition-aligned VLA model that employs instruction-driven routing and visual token sparsification to retain task-relevant information and improve cross-modal coordination.

\end{itemize}

\section{Ablation Studies}
\label{app:Ablation}

\paragraph{Evaluation Setup.}
We conduct ablation studies on RoboCasa to examine the contributions of key components in CMC-WAM. All configurations are evaluated on the same 24 tasks, with 50 valid episodes per task and matched environment initializations. We consider five ablation variants to investigate the effects of prediction-guided action selection, shared/private factorization, independent mechanism routing, and asymmetric coupling.

\paragraph{Ablation Configurations.}  

\textbf{(1) Cosmos + W$\rightarrow$A (no CMC).}
We add prediction-guided candidate selection to the original Cosmos Policy. The pretrained world prediction model and value function evaluate four candidate actions based on their predicted future states. The candidate with the highest value is selected for execution. No CMC factorization, routing, or expert modules are introduced.

\textbf{(2) w/o Factorization.}
We remove the explicit shared/private factorization and directly use the unfactorized Action and World features for subsequent routing and expert modeling. Since the feature dimensions change, we adapt them to match the input dimensions of the original routers and experts. We also remove the corresponding shared consistency loss $L_{\mathrm{share}}$ and redundancy suppression loss $L_{\mathrm{red}}$. Independent routers, experts, and both coupling mechanisms are retained.

\textbf{(3) Shared Router.}
We replace the two independent Action and World routers with a single shared-parameter router. The branch-specific conditioning inputs and separate expert sets are retained, together with shared/private factorization and both coupling mechanisms.

\textbf{(4) w/o A$\rightarrow$W Coupling.}
We remove the additional action conditioning introduced by CMC from both the World Router and World Experts, while retaining the original action-conditioned world prediction in Cosmos. All other CMC components and W$\rightarrow$A prediction-guided selection remain unchanged.

\textbf{(5) w/o W$\rightarrow$A Coupling.}
We disable multi-candidate selection based on predicted future states by setting the number of candidates to $N=1$. The generated action sequence is executed directly, without value-based candidate ranking. Shared/private factorization, independent routing, expert networks, and A$\rightarrow$W coupling are retained.

\paragraph{Training and Inference Protocol.}
The architectural variants, including w/o Factorization, Shared Router, and w/o A$\rightarrow$W Coupling, are retrained using their respective modified architectures. They share the same pretrained Cosmos-Predict2.5-2B backbone, training data, optimizer, and number of training steps.

For Cosmos + W$\rightarrow$A (no CMC), we directly use the original pretrained Cosmos model, including its world prediction model and value function, without additional CMC training. For w/o W$\rightarrow$A Coupling, we use the trained Full CMC-WAM checkpoint and disable multi-candidate selection only during closed-loop inference by setting $N=1$.

\paragraph{Results.}
As reported in Table~\ref{tab:ablation}, adding W$\rightarrow$A prediction-guided selection alone improves the original Cosmos Policy by only 0.1 percentage points. Full CMC-WAM achieves a further gain of 2.4 percentage points over this variant. Removing shared/private factorization, replacing independent routers with a shared router, and removing A$\rightarrow$W or W$\rightarrow$A coupling reduce the success rate by 1.8, 1.2, 1.3, and 1.9 percentage points, respectively. These results demonstrate the contributions of structured factorization, independent mechanism routing, and asymmetric coupling to closed-loop performance.

\section{Detailed Mechanism Analysis}
\label{app:mechanism_analysis}

We analyze CMC-WAM from three perspectives: factor specialization, dynamic mechanism routing, and asymmetric mechanism coupling. These analyses examine the predictive roles of shared and private factors, the stage-dependent expert selection of the two mechanism routers, and the information exchange between action generation and world prediction.

\paragraph{Factor Specialization.}
We evaluate the predictive roles of shared and private factors by examining their ability to predict common interaction states, actions, and future states. We employ Linear Probes with identical architectures to assess the predictive information encoded in different factors. For action and future-state prediction, we calculate the $R^2$ gains of the corresponding private factors over the shared factor under the same prediction tasks.

As reported in Table~\ref{tab:mechanism_analysis}, the shared factor achieves an $R^2$ of 0.70 for predicting common interaction states. The Action Private factor yields an $R^2$ gain of 0.13 over the shared factor on action prediction, while the World Private factor achieves a gain of 0.16 on future-state prediction. These results suggest that the shared factor captures common interaction information, while the private factors provide complementary information for their respective generative processes.

\paragraph{Dynamic Mechanism Routing.}
We examine how the two mechanism routers adapt their expert selection across interaction stages. Specifically, we aggregate the normalized Top-2 routing weights within each stage to obtain stage-level routing distributions. We then measure the distributional differences between stages using the average pairwise Jensen--Shannon (JS) divergence:

\begin{equation}
D_{\mathrm{phase}}^m =
\frac{2}{K(K-1)}
\sum_{i<j}
\operatorname{JS}
\left(P_m^{(i)},P_m^{(j)}\right),
\end{equation}

where $m\in\{A,W\}$ denotes the Action or World Router, $K$ is the number of interaction stages, and $P_m^{(i)}$ represents the aggregated routing distribution of mechanism $m$ in stage $i$.

As reported in Table~\ref{tab:mechanism_analysis}, the Action and World routers achieve average pairwise JS divergences of 0.25 and 0.32, respectively. Both routers exhibit changes in their expert-selection distributions across interaction stages, with the World Router showing greater distributional variation. These results indicate that expert selection adapts to changing interaction conditions in both generative processes.

\paragraph{Asymmetric Mechanism Coupling.}
We further evaluate the information exchange between action generation and world prediction in both directions: Action-to-World (A$\rightarrow$W) and World-to-Action (W$\rightarrow$A).

For A$\rightarrow$W coupling, we examine the effect of action conditions on future-state prediction. We compare matched and shuffled action conditions. The matched condition uses the action sequence corresponding to the current interaction trajectory, while the shuffled condition uses a permuted action sequence. We measure the prediction error using the mean squared error (MSE) between predicted and ground-truth future latent states. The error reduction is defined as

\begin{equation}
\Delta E_{\mathrm{action}}
=
E_{\mathrm{shuffled}}
-
E_{\mathrm{matched}}.
\end{equation}

The matched action condition reduces future-state prediction error by 0.049 compared with the shuffled condition. This result indicates that action-aligned conditioning provides useful information for future-state prediction.

For W$\rightarrow$A coupling, we examine whether predicted future states can support candidate-action evaluation. For each candidate action sequence, the World Model predicts its corresponding future state, which is then evaluated by the pretrained task-related value function:

\begin{equation}
J^{(n)}
=
V_G\left(\hat{s}_{t+H}^{(n)},G\right),
\end{equation}

where $\hat{s}_{t+H}^{(n)}$ denotes the predicted future state of the $n$-th candidate, and $G$ represents the task goal.

For evaluation, we sample decision states from RoboCasa and generate four candidate action sequences for each state. The candidates are ranked according to their predicted values. To obtain an independent reference, we restore the simulator to the same state and execute each candidate separately. Their relative performance is determined from the observed task outcomes, independently of the predicted future states and value scores. Each set of four candidates yields up to six pairwise comparisons. Pairwise Ranking Accuracy measures the proportion of valid candidate pairs for which the predicted and execution-based preferences agree.

The resulting Pairwise Ranking Accuracy reaches 78.8\%, indicating that predicted future states provide useful information for assessing the relative merits of candidate actions. Overall, these analyses provide complementary evidence for factor specialization, dynamic mechanism routing, and asymmetric mechanism coupling. The results reveal differentiated predictive roles of shared and private factors, stage-dependent expert selection, and complementary information exchange between action generation and world prediction.

\begin{table}[t]
    \centering
    \caption{
        \textbf{Quantitative mechanism analysis of CMC-WAM.}
        We evaluate factor specialization, dynamic mechanism
        routing, and asymmetric coupling between action
        generation and world prediction.
    }
    \label{tab:mechanism_analysis}
    \small
    \setlength{\tabcolsep}{5pt}
    \renewcommand{\arraystretch}{1.15}
    \begin{tabular}{lcc}
        \toprule
        Analysis & Metric & Result \\
        \midrule
        \multirow{2}{*}{Factor Specialization}
        & $R^2_{\mathrm{common}}$ & 0.70 \\
        & $\Delta R_A^2 / \Delta R_W^2$ & +0.13 / +0.16 \\
        \midrule
        Dynamic Routing
        & $D_{\mathrm{phase}}^A / D_{\mathrm{phase}}^W$
        & 0.25 / 0.32 \\
        \midrule
        A$\rightarrow$W Coupling
        & $\Delta E_{\mathrm{action}}$ & +0.049 \\
        W$\rightarrow$A Coupling
        & Pairwise Ranking Accuracy & 78.8\% \\
        \bottomrule
    \end{tabular}
\end{table}

% \begin{table*}[t]
%     \centering
%     \caption{
%         \textbf{Generalization and efficiency analysis of CMC-WAM.}
%         We evaluate robustness under shifted initial configurations and analyze the trade-off between candidate number and inference cost.
%     }
%     \label{tab:generalization_efficiency}
%     \small
%     \setlength{\tabcolsep}{9pt}
%     \renewcommand{\arraystretch}{1.15}
%     \begin{tabular*}{\textwidth}{@{\extracolsep{\fill}}lccc@{}}
%         \toprule
%         \multicolumn{4}{l}{\textit{(a) Generalization under distribution shifts}} \\
%         \midrule
%         Evaluation Setting & Cosmos Policy & CMC-WAM & $\Delta$ SR \\
%         \midrule
%         In-distribution
%             & 71.5 & 74.2 & +2.7 \\
%         Shifted Initial Configurations
%             & 61.7 & 66.7 & +5.0 \\
%         \midrule
%         \multicolumn{4}{l}{\textit{(b) Candidate number and inference efficiency}} \\
%         \midrule
%         Metric & $N=1$ & $N=4$ & Change \\
%         \midrule
%         Success Rate (\%)
%             & 72.1 & 74.0 & +1.9 \\
%         Latency (ms/decision)
%             & 227 & 617 & $2.72\times$ \\
%         \midrule
%         \multicolumn{4}{l}{\textit{(c) Model overhead}} \\
%         \midrule
%         Additional Trainable Parameters
%             & \multicolumn{3}{c}{0.35M (0.018\% of the backbone)} \\
%         Inference Overhead ($N=4$ vs. Cosmos)
%             & \multicolumn{3}{c}{+390 ms/decision (+172\%)} \\
%         \bottomrule
%     \end{tabular*}
% \end{table*}
% \section{Generalization and Efficiency Analysis}
% \label{app:Generalization}

\begin{table*}[t]
    \centering
    \caption{
        \textbf{Generalization and efficiency analysis of CMC-WAM.}
        We evaluate robustness under shifted initial configurations
        and analyze the trade-off between candidate number and
        inference cost. In-distribution results are reported as
        mean $\pm$ sample standard deviation over three independent
        runs. Shifted-configuration and candidate-number comparisons
        are obtained from separate single-run paired evaluations.
    }
    \label{tab:generalization_efficiency}
    \small
    \setlength{\tabcolsep}{9pt}
    \renewcommand{\arraystretch}{1.15}
    \begin{tabular*}{\textwidth}
        {@{\extracolsep{\fill}}lccc@{}}
        \toprule

        \multicolumn{4}{l}
        {\textit{(a) Generalization under initial-configuration shifts}} \\
        \midrule
        Evaluation Setting
            & Cosmos Policy
            & CMC-WAM
            & $\Delta$ SR (pp) \\
        \midrule
        In-distribution (3 runs)
            & $71.5 \pm 0.2$
            & $74.2 \pm 0.3$
            & $+2.7$ \\

        Shifted Initial Configurations (1 run)
            & $61.7$
            & $66.7$
            & $+5.0$ \\

        \midrule
        \multicolumn{4}{l}
        {\textit{(b) Candidate number and inference efficiency}} \\
        \midrule
        Metric
            & $N=1$
            & $N=4$
            & Change \\
        \midrule
        Success Rate (\%)
            & $72.1$
            & $74.0$
            & $+1.9$ pp \\

        Latency (ms/decision)
            & $227$
            & $617$
            & $2.72\times$ \\

        \midrule
        \multicolumn{4}{l}
        {\textit{(c) Parameter and inference overhead}} \\
        \midrule
        Additional Trainable Parameters
            & \multicolumn{3}{c}
            {$0.35$M ($0.018\%$ of the 2B backbone)} \\

        Additional Inference Latency ($N=4$ vs. $N=1$)
            & \multicolumn{3}{c}
            {$+390$ ms/decision ($+172\%$)} \\

        \bottomrule
    \end{tabular*}
\end{table*}

\section{Generalization and Efficiency Analysis}
\label{app:Generalization}

We further evaluate the generalization and computational efficiency of CMC-WAM under different interaction conditions. Specifically, we examine its robustness to initial-configuration shifts, the effect of candidate number on prediction-guided planning, and the additional model and inference overhead.

\paragraph{Robustness to Initial-Configuration Shifts.}
To evaluate robustness beyond the standard initialization distribution, we conduct an additional evaluation using held-out initial configurations on RoboCasa. We introduce controlled perturbations to task-relevant initial object poses and articulated-object states while keeping the task semantics and scene configurations unchanged. For rigid-object manipulation, we perturb the initial XY positions of manipulated objects by 5--10 cm. For orientation-sensitive objects, we additionally apply yaw perturbations of 15--30 degrees. For articulated-object manipulation, we perturb the initial joint states within 10\%--20\% of their valid motion ranges. The shifted configurations are sampled from regions that do not overlap with the nominal initialization regions. All perturbed states undergo collision, workspace, and physical validity checks to ensure feasible initial configurations.

The task instructions, object categories, target configurations, scene geometry, robot initial joint configurations, camera settings, and physics parameters remain unchanged. Both methods are evaluated on identical shifted initializations. Following the standard RoboCasa protocol, we evaluate 50 valid episodes per task across all 24 tasks, resulting in 1,200 episodes for each method. As shown in Table~\ref{tab:generalization_efficiency}(a), CMC-WAM achieves a success rate of 66.7\%, compared with 61.7\% for Cosmos Policy, yielding a 5.0 percentage-point improvement. This additional evaluation shows that CMC-WAM retains a performance advantage under initial-configuration shifts. These results are reported separately from the three-run in-distribution results, as they come from an additional single evaluation run.

\paragraph{Effect of Candidate Number.}
We further investigate the effect of candidate number on prediction-guided planning. We compare single-candidate execution ($N=1$) with four-candidate selection ($N=4$), using the same model weights and matched environment initializations. For each candidate, the World branch predicts the corresponding future state. The pretrained Cosmos Value Function then evaluates the predicted state, and the candidate with the highest value is selected for execution. Each candidate predicts 32 future steps, of which the first 16 steps are executed before replanning with updated observations. As reported in Table~\ref{tab:generalization_efficiency}(b), increasing the candidate number from one to four improves the success rate from 72.1\% to 74.0\%, corresponding to a 1.9 percentage-point gain. This comparison uses a single evaluation run with matched environment initializations. Meanwhile, inference latency increases from 227 ms to 617 ms per decision, corresponding to a 2.72$\times$ increase. These results show that evaluating multiple predicted futures can improve action selection, at the cost of additional online computation.

\paragraph{Parameter and Inference Overhead.}
CMC-WAM introduces only 0.35M additional trainable parameters, accounting for approximately 0.018\% of the 2B backbone parameters. The pretrained backbone and generation heads remain frozen. Training is restricted to the newly introduced CMC modules, including the projections, routers, and experts. Regarding online inference, four-candidate W$\rightarrow$A planning increases the latency by 390 ms per decision compared with single-candidate execution ($N=1$). This corresponds to a 172\% increase. The reported latency covers the complete model decision process, including candidate generation, world prediction, value evaluation, and candidate selection. Overall, CMC-WAM achieves structured mechanism modeling with a small increase in trainable parameters. Its prediction-guided planning further improves closed-loop performance, with additional computational cost during online inference.

\begin{table}[t]
    \centering
    \caption{
        \textbf{Real-world manipulation results.}
        SR across three physical manipulation tasks, with 30 independent trials per task.
    }
    \label{tab:real_world}
    \small
    \setlength{\tabcolsep}{3pt}
    \renewcommand{\arraystretch}{1.15}
    \begin{tabular*}{\columnwidth}{@{\extracolsep{\fill}}p{0.43\columnwidth}p{0.30\columnwidth}c@{}}
        \toprule
        Task & Interaction Type & Success / SR \\
        \midrule
        Cup Handover and Target Placement
            & Bimanual Transfer & 26/30 (86.7\%) \\
        Plate Insertion into Rack
            & Constrained Insertion & 23/30 (76.7\%) \\
        Remote Control Placement into Container
            & In-Container Placement & 27/30 (90.0\%) \\
        \midrule
        \textbf{Overall}
            & -- & \textbf{76/90 (84.4\%)} \\
        \bottomrule
    \end{tabular*}
\end{table}

\section{Real-World Experimental Details}
\label{app:real_world}

To further evaluate CMC-WAM in physical environments, we deploy it on a dual-arm platform consisting of two 7-DoF NERO robotic arms from AgileX Robotics, each equipped with a parallel gripper. CMC-WAM uses the observation and action spaces of the physical platform while retaining its mechanism factorization, independent routing, and prediction-guided action selection. The policy operates in a closed loop using visual observations, robot states, and language task instructions.

We evaluate three tabletop manipulation tasks. \textbf{Cup Handover and Target Placement} requires transferring a cup between the two arms and placing it at a designated target position. \textbf{Plate Insertion into Rack} involves grasping a plate, adjusting its orientation, and inserting it into a rack. \textbf{Remote Control Placement into Container} requires picking up a remote control and placing it inside a target container. These tasks cover bimanual transfer, constrained insertion, and in-container placement.

For each task, the robot starts from a predefined initial configuration, while object positions and orientations are varied within a predefined workspace. We conduct 30 independent trials per task, resulting in 90 real-world trials. A trial is considered successful only when the robot completes the instructed manipulation and reaches the target state without manual intervention or correction. We use task success rate as the primary evaluation metric.

The results are summarized in Table~\ref{tab:real_world}, with representative execution sequences shown in Figure~\ref{fig:real_world_tasks}. The experiments demonstrate the applicability of CMC-WAM to physical manipulation tasks with different interaction requirements.

\section{Limitations and Future Directions}
\label{app:limitations}

\paragraph{Mechanism Representation.}
CMC-WAM explicitly models the distinct structural dependencies of action generation and world evolution through shared/private factorization and independent mechanism routing. However, the learned representations do not necessarily recover the underlying causal mechanisms of physical interactions. Further investigation into their interpretability and stability across different interaction conditions remains an important direction.

\paragraph{Prediction-Guided Planning.}
The W$\rightarrow$A mechanism relies on the accuracy of future-world predictions and task-value estimates. Prediction errors may affect candidate ranking and action selection. Moreover, evaluating multiple predicted futures increases online inference cost. Future work may explore uncertainty-aware evaluation and adaptive candidate selection to improve decision reliability and computational efficiency.

\paragraph{Generalization.}
Our experiments primarily focus on Cosmos-Predict2.5, with evaluations on RoboCasa, LIBERO, and a limited set of real-world manipulation tasks. Although CMC-WAM shows improved robustness under initial-configuration shifts, its applicability to other WAM architectures, robotic platforms, and more diverse physical environments remains to be investigated. Extending mechanism modeling across architectures and interaction settings is a promising direction for future research.

% WARNING: do not forget to delete the supplementary pages from your submission 
% \input{sec/X_suppl}

\end{document}